\documentclass[11pt]{article}

\usepackage{lineno}

\usepackage{lscape}
\usepackage{graphicx}
\usepackage{amsmath}
\usepackage{amsthm}
\usepackage{amssymb}
\usepackage{booktabs}
\usepackage[frozencache=true]{minted}
\usepackage{enumitem}

\usepackage{stmaryrd}

\usepackage[hidelinks]{hyperref}

\usepackage{todonotes}
\usepackage{xcolor}
\usepackage{standalone}
\usepackage{pgfplots}
\pgfplotsset{compat=1.18}

\usepackage{tikz, pgf}
\usetikzlibrary{arrows, automata, positioning, calc, fit, math, matrix, decorations.pathreplacing}

\usepackage[linesnumbered]{algorithm2e}
\usepackage[nolist]{acronym}

\usepackage[numbers]{natbib}

\usepackage{svg}

\usepackage{marvosym}
\usepackage{tabularx}

\usepackage{float}
\usepackage{multirow}

\usepackage{xspace}
\usepackage{soul}

\usepackage{ulem}

\usepackage{bbm}
\usepackage{nicefrac}
\usepackage{placeins}

\acrodef{NN}{neural network}
\acrodefplural{NN}[NNs]{neural networks}

\acrodef{NLDT}[NDT]{neurosymbolic decision tree}

\acrodef{PDT}{probabilistic decision tree}
\acrodefplural{PDT}[PDTs]{probabilistic decision trees}

\acrodef{SDT}{soft decision tree}

\acrodef{BDT}{Boolean decision tree}

\acrodef{DT}{decision tree}
\acrodefplural{DT}[DTs]{decision trees}

\acrodef{MLP}{multilayer perceptron}

\acrodef{IG}{Information Gain}

\acrodef{NeSy}{neurosymbolic}
\acrodef{ML}{Machine Learning}
\acrodef{MR}{Machine Reasoning}

\acrodef{CWA}{closed world assumption}
\acrodef{LP}{logic programming}

\acrodef{DAG}{directed acyclic graph}

\acrodef{TDIDT}{top-down induced Decision Trees}

\acrodef{AC}{arithmetic circuit}
\acrodef{WMC}{weighted model counting}

\acrodef{KC}{knowledge compilation}

\acrodef{CE}{cross-entropy}

\newlist{mylist}{enumerate*}{1}
\setlist[mylist]{label=(\roman*)}

\title{Neurosymbolic Decision Trees}

\author{
\small Matthias Möller\textsuperscript{1}, Arvid Norlander, Pedro Zuidberg Dos Martires\textsuperscript{1},\\\small Luc De Raedt\textsuperscript{1,2}\\[0.5em]
\footnotesize \textit{\textsuperscript{1}AASS, Örebro University Sweden}\\
\footnotesize \textit{\textsuperscript{2}Department of Computer Science, Leuven.AI, KU Leuven Belgium}
}
\date{}

\newcommand{\eg}{e.g.}
\newcommand{\ie}{i.e.}

\newcommand{\namealgo}{NeuID3}

\newcommand{\graphicsdir}{./graphics}

\definecolor{classtruecolor}{HTML}{3FBC9D}
\definecolor{classfalsecolor}{HTML}{EE2967}

\newcommand{\checked}{\textcolor{classtruecolor}{\Checkedbox}}
\newcommand{\unchecked}{\textcolor{classfalsecolor}{\CrossedBox}}

\setminted[problog.py:ProbLogLexer -x]{
xleftmargin=0pt,
linenos=false,
escapeinside=@@,
mathescape=true,
breaklines=true,
fontfamily=courier,
breakautoindent=True
fontsize=\small
}

\newcounter{lstlabelcounter}

\newminted[problog]{problog.py:ProbLogLexer -x}{}

\setminted[text]{
xleftmargin=20pt,
linenos=false,
mathescape=true,
escapeinside=@@,
breaklines=true,
}
\newminted[dcplp]{text}{}

\renewcommand{\impliedby}{\leftarrow}

\newcommand{\test}{\tau}

\DeclareMathOperator*{\probcoco}{\mathtt{{:}{:}}}

\DeclareMathOperator*{\argmax}{arg\,max}

\begin{document}


    \tikzstyle{sample}=[
    shape=rectangle,
    draw,
    align=left,
    minimum height=1.7cm
    ]

    \tikzstyle{code}=[
    shape=rectangle,
    draw,
    align=left,
    minimum height=3.1cm
    ]

    \tikzstyle{leafnode}=[
    shape=rectangle,
    fill=white,
    draw, align=center,
    ]

    \tikzstyle{inode}=[
    fill=white,
    shape=rectangle,
    draw,
    rounded corners,
    align=center
    ]

    \tikzstyle{tree-distance} = [
    level 1/.style={sibling distance=15em},
    level 2/.style={sibling distance=7.5em},
    level distance=5em
    ]

    \tikzstyle{marked}=[
    draw=forest
    ]

    \tikzstyle{prob-distribution}=[%
            anchor=center,
            ybar,
            width=2cm,
            height=2cm,
            xtick=\empty,
            ymax=1.0
    ]

\maketitle

\begin{abstract}
 
 \Ac{NeSy} AI studies the integration of \acp{NN} and symbolic reasoning based on logic.
 Usually,
 \ac{NeSy} techniques focus on learning the neural, probabilistic and/or fuzzy parameters of \ac{NeSy} models. 
Learning the symbolic or logical structure of such models has, so far, received less attention. We introduce \acp{NLDT}, as an extension of decision trees together with a novel \ac{NeSy} structure learning algorithm, which we dub \namealgo{}.
\namealgo{} adapts the standard top-down induction  of decision tree algorithms and combines it with
a neural probabilistic logic representation, inherited from the DeepProbLog family of models. 
The key advantage of learning \acp{NLDT} with \namealgo{} is the support of both symbolic and subsymbolic data (such as images), and that they can exploit background knowledge during the induction of the tree structure,
In our experimental evaluation we demonstrate the benefits of \acp{NeSy} structure learning over more traditonal approaches such as purely data-driven learning with neural networks.

\end{abstract}

\section{Introduction}
\acresetall

\Ac{NeSy} AI combines symbolic reasoning methods and \acp{NN} in order to leverage their respective strengths: the ability to exploit background knowledge using logic and the function approximation capabilities of \acp{NN}~\cite{manhaeveNeuralProbabilisticLogic2021, greffBindingProblemArtificial2020, yuSurveyNeuralsymbolicLearning2023}.
A common approach to constructing \ac{NeSy} systems starts from probabilistic logic formulas, where, in contrast to deterministic logic formulas, the literals in a formula are true or false with a certain probability.
By predicting the probabilities using \acp{NN} we obtain a \ac{NeSy} model.
However, this approach requires the logical structure of the \ac{NeSy} model to be given upfront, 
which is not possible in many practical settings.   This paper addresses the challenging problem
of simultaneously learning both the structure and the parameters for \ac{NeSy}. 
More specifically,
\begin{enumerate}[itemsep=0pt, parsep=0pt, topsep=0pt]
    \item we introduce \acp{NLDT} by parametrizing probabilistic decision trees using \acp{NN} (\autoref{sec:NLDT}),
    \item we present \namealgo{}\footnote{The name Neural ID3 (NeuID3) is chosen after Ross J. Quinlan's famous ID3 (Inductive Dichotomizer 3) algorithm \cite{quinlanInductionDecisionTrees1986}.} an iterative top-down decision tree learner for inducing \acp{NLDT} over tree-shaped  logic formulas (\autoref{sec:LearningNDTs}).
\end{enumerate}

In our experimental evaluation (\autoref{sec:experiments}) we then demonstrate that \acp{NLDT} can use symbolic as well as subsymbolic features in order to classify data points. Compared to standard decision trees, this has the advantage of being able to use raw sensor data, \eg{} images, as features. Compared to \acp{NN}, \acp{NLDT} have the advantage of allowing the incorporation of background knowledge, leading, for instance, to more data-efficient learners.
Compared to other NeSy approaches, NDTs have the advantage that the symbolic structure of the model is completely induced from scratch. We discuss the differences to related \ac{NeSy} approaches in \autoref{sec:related_work}.

Note that throughout the paper, we focus largely on propositional logic for simplicity,
though the introduced techniques can in principle be generalized towards first order logic following the works of \cite{blockeelTopinductionFirstorderLogical1998, deraedtProbLogProbabilisticProlog2007,manhaeveNeuralProbabilisticLogic2021}.

\section{Background}
\label{sec:background}

We first introduce the necessary concepts and notation for propositional logic programming. Then we describe Boolean decision trees and their probabilistic extension in terms of (probabilistic) logic programming \cite{de2015probabilistic}.

An atom is a proposition $q$ (sometimes also called a Boolean variable), and a literal is an atom or its negation $\neg q$. 
 A rule is an expression of the form $h \leftarrow b_1 \wedge ... \wedge b_n$ where $h$ is an atom and the $b_i$ are literals. The meaning
of such a rule is that $h$ holds whenever the conjunction of all $b_i$ holds. Rules with an empty body $n = 0$ are called facts and hold unconditionally.

A set of rules and facts constitutes a logic program $L$ and can be used to decide
whether a literal $q$ logically follows from the program. We write $L \models q$ when it does ($L$ models $q$), and $L \not\models q$ when it does not. 

\subsection{Boolean Decision Trees}
\label{ssec:BCDT}

\acp{BDT}
can be represented as a binary tree, cf. \autoref{fig:structureDT}. Each internal node $i$ (including the root) contains a proposition $\test_i$, also called a \textit{test}. 
The two children of an internal node then represent 
the two possible value assignment $\test_i = true$ and $\test_i =false$, respectively. 
We will use $\cal F$ to refer to the set of all propositions used as tests in the decision tree. For the decision tree in \autoref{fig:structureDT}, we are given $\mathcal{F} = \{ burglary, earthquake, alarm\}$.

By associating each leaf with one of two classes, which we denote by $pos$ and $neg$, respectively,
we can use \Acp{BDT} as binary classification decision trees. For the BDT in \autoref{fig:structureDT} we can pick for instance the following leaf-to-class mapping (written as rules):
\begin{align*}
  pos &\impliedby leaf_1.\quad pos \impliedby leaf_3.\\
  neg &\impliedby leaf_2.\quad neg \impliedby leaf_4.\quad neg \impliedby leaf_5.
\end{align*}
To classify an example with a BDT one starts evaluating the atom in the root, and propagates the example towards the child corresponding to the truth-value of the atom. This process is repeated recursively until a leaf node is reached and the class assigned to that leaf node is predicted for the example.

As an illustrative example, consider the two sets of literals $\mathcal{L}_{e_1}$ and $\mathcal{L}_{e_2}$ representing two examples $e_1$ and $e_2$
\begin{align*}
  \mathcal{L}_{e_1} &= \{\phantom{\neg} burglary, \neg earthquake, \phantom{\neg} alarm\}\\
  \mathcal{L}_{e_2} &= \{\neg burglary, \neg earthquake, \neg alarm\}
\end{align*}
where a positive literal indicates that the test is assigned with true while a negated ($\neg$) literal indicates that the test evaluates to false.
The corresponding subsets of the positive literals (atoms) are $\mathcal{F}_{e_1} = \{burglary, alarm\}$ and $\mathcal{F}_{e_2} = \{ \}$, respectively.
In the context of the \ac{BDT} in \autoref{fig:structureDT}, $e_1$ represents the situation where an $alarm$ is triggered by a $burglary$. In contrast, $e_2$ represents the case of no $alarm$ being triggered and the absence of a $burglary$ and no $earthquake$.

We can also take a logic programming perspective on classification with \acp{BDT}, as it is well-known that decision trees can be transformed into a set of
logical rules~\cite{mitchellMachineLearning2013}. We illustrate this on on the right in \autoref{fig:structureDT}.
Classification then proceeds as follows: 
an example $e$ is labeled with class $c\in \{pos, neg\}$ if and only if $L \cup \mathcal{F}_e \models c$, where $\mathcal{F}_e$ is the subset of positive literals (atoms) in $\mathcal{L}_e$. The choice for using $\mathcal{F}_e$ results from the principle of \textit{negation as a failure} which states that all atoms that cannot be proven true are assumed to be false~\cite{flachSimplyLogicalIntelligent1994}.
Reconsidering the set of literals $\mathcal{L}_{e_1}$ and $\mathcal{L}_{e_1}$ we consequently have $L \cup \mathcal{F}_{e_1} \models pos$ and $L \cup \mathcal{F}_{e_2} \models neg$.

\begin{figure}
  \resizebox{\columnwidth}{!}{
    \ifstandalone
    \usepackage{tikz}
    \usepackage{xcolor}
    \usetikzlibrary{positioning, calc, fit, math, matrix, arrows}
    \include{./tree_style.tex}

    \begin{document}
\fi

\begin{tikzpicture}
    \tikzstyle{tree-distance}=[
    level 1/.style={sibling distance=20em},
    level 2/.style={sibling distance=8.5em},
    level distance=5em
    ]

    \coordinate(column-1) at (0,0);
    \coordinate(column-2) at (10.5,0);
    \coordinate(column-3) at (16.5,0);
    \coordinate(row-1) at (0,0);
    \coordinate(row-2) at (0,-6.5);

    \pgfdeclarelayer{foreground}
    \pgfdeclarelayer{background}
    \pgfsetlayers{background, foreground}

    {
        \newcommand{\rootname}{root}

        \begin{scope}
            \begin{scope}[tree-distance, edge from parent/.style={}]
                \node (\rootname) at (row-1 -| column-1) {}
                child {child child}
                child{child{child child} child};
            \end{scope}

            \begin{pgfonlayer}{foreground}
                \node[inode] (\rootname) at (\rootname) {$burglary$};
                \node[inode] (\rootname-1) at (\rootname-1) {$alarm$};
                
                \node[leafnode] (\rootname-1-1) at (\rootname-1-1) {$leaf_1$};
                
                \node[leafnode] (\rootname-1-2) at (\rootname-1-2) {$leaf_2$};
                
                \node[inode]    (\rootname-2) at (\rootname-2) {$earthquake$};
                \node[inode] (\rootname-2-1) at (\rootname-2-1) {$alarm$};
                
                \node[leafnode] (\rootname-2-1-1) at (\rootname-2-1-1) {$leaf_3$};
                
                \node[leafnode] (\rootname-2-1-2) at (\rootname-2-1-2) {$leaf_4$};

                \node[leafnode] (\rootname-2-2) at (\rootname-2-2) {$leaf_5$};                
                
            \end{pgfonlayer}

            \begin{pgfonlayer}{background}
                \begin{scope}[->]
                    \draw (\rootname) edge[midway, above left] node {true} (\rootname-1);
                    \draw (\rootname-1) edge[midway, above left] node {true} (\rootname-1-1);
                    \draw (\rootname-1) edge[midway, above right] node {false} (\rootname-1-2);

                    \draw (\rootname) edge[midway, above right] node {false} (\rootname-2);
                    \draw (\rootname-2) edge[midway, above left] node {true} (\rootname-2-1);
                    \draw (\rootname-2-1) edge[midway, above left] node {true} (\rootname-2-1-1);
                    \draw (\rootname-2-1) edge[midway, above right] node {false} (\rootname-2-1-2);
                    \draw (\rootname-2) edge[midway, above right] node {false} (\rootname-2-2);
                    \node[fit= (\rootname) (\rootname-1-1) (\rootname-2-2)] (\rootname-all) {};
                \end{scope}
            \end{pgfonlayer}
        \end{scope}

         \begin{scope}

             \begin{pgfonlayer}{foreground}
            \node (header) at ($(row-1 -| column-2) + (0,0)$) {\textbf{Rules}};
            \node[anchor=north] (code) at ($(header.south) + (0,-0.1)$) {
                \begin{minipage}{0.6\textwidth}
                \begin{align*}
leaf_1 &\impliedby burglary \land alarm.\\
leaf_2 &\impliedby burglary \land \neg alarm.\\
leaf_3 &\impliedby \neg burglary \land earthquake \land alarm.\\
leaf_4 &\impliedby \neg burglary \land earthquake \land \neg alarm.\\
leaf_5 &\impliedby \neg burglary \land \neg earthquake.\\
                \end{align*}
                \end{minipage}
            };
             \end{pgfonlayer}

         \end{scope}




            

    }

\end{tikzpicture}

\ifstandalone
    \end{document}
\fi
  }
  \caption{\label{fig:structureDT} 
    Left: A \ac{BDT} modelling whether a person calls the police conditioned on whether a burglary, an alarm, and an earthquake occur.
    Right: a set of rules representing the \ac{BDT} as a logic program. 
  }

\end{figure}
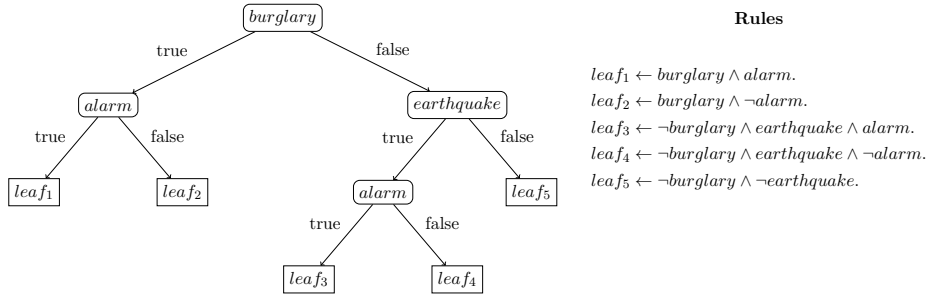

\subsection{Probabilistic Boolean Decision Tree}
\label{ssec:PDT}

\begin{figure}[!b]
  \resizebox{\textwidth}{!}{
    \def\mathunderline#1#2{\color{#1}\underline{{\color{black}#2}}\color{black}}

\begin{tikzpicture}
    \coordinate(column-1) at (0,0);
    \coordinate(column-2) at (12,0);
    \coordinate(column-3) at (17,0);
    \coordinate(row-2) at (0,-7.5);
    \coordinate(row-1) at (0,0);
    \coordinate(row-0) at (0,5.);

    \pgfplotsset{set layers,
        every axis/.style={
            anchor=center,
            ybar,
            width=2.5cm,
            height=2.0cm,
            xtick=\empty,
            ymax=1.0,
            ymin=0.0,
            bar shift=0pt,
            bar width=0.25cm,
            xmin=0,
            xmax=3,
            x tick label style={
                rotate=45,anchor=north east, font=\tiny
            },
            y tick label style={
                font=\tiny,
                xshift = -0.2em
            },
            y tick style = {draw=none},
            ymajorgrids=true
    }}

    \pgfdeclarelayer{foreground}
    \pgfdeclarelayer{background}
    \pgfsetlayers{background, foreground, axis grid, axis ticks, axis lines, axis descriptions, axis tick labels}

    \newcommand{\drawbetween}[4]{
        \coordinate (help) at ($(#1)!0.5!(#2)$);
        \coordinate (help2) at (#2 |- #1);
        \node[align=center] at ($(help)!#4!(help2)$) {#3};
    }

    \newcommand{\probist}[3]{%

        \begin{axis}[at={(#1)}, name=#1-axis,
            ytick={#2, #3},
            yticklabels={%
            \textcolor{classtruecolor}{\small #2},%
            },
            extra y ticks={#3}, 
            extra y tick labels={\textcolor{classfalsecolor}{\footnotesize #3}}, 
            extra y tick style={
                ticklabel pos=right, 
                yticklabel style={anchor=west}, 
                xshift=5pt,
            },
            grid=none,
            yticklabel style={anchor=east}
            ]
            \addplot+[classtruecolor, mark=none] coordinates {
                (1, #2)
            };
            \addplot+[classfalsecolor, mark=none] coordinates {
                (2, #3)
            };
        \end{axis}

        \begin{pgfonlayer}{background}
            \node[leafnode, fit=(#1-axis)] (#1) {};
        \end{pgfonlayer}
    }

    {
        \tikzstyle{tree-distance} = [
        level 1/.style={sibling distance=20em},
        level 2/.style={sibling distance=9em},
        level distance=5em
        ]

        \newcommand{\rootname}{root3}
        \begin{scope}

            \begin{scope}[tree-distance, edge from parent/.style={}]
                \node (\rootname) at (row-1 -| column-1) {}
                child{child child}
                child {child {child child} child};
            \end{scope}

            \begin{pgfonlayer}{foreground}
                \node[inode] (\rootname) at (\rootname) {$burglary$};
                \node[inode] (\rootname-1) at (\rootname-1) {$alarm$};

                \node[inode] (\rootname-2) at (\rootname-2) {$earthquake$};

                \node[inode] (\rootname-2-1) at (\rootname-2-1) {$alarm$};

                \probist{\rootname-1-1}{0.95}{0.05}
                \node[anchor=south] at ($(\rootname-1-1.north)+(-0.1,0)$) {$leaf_1$};
                \probist{\rootname-1-2}{0.0}{1.0}
                \node[anchor=south] at ($(\rootname-1-2.north)+(0.1,0)$) {$leaf_2$};
                \probist{\rootname-2-2}{0.001}{0.999}
                \node[anchor=south] at ($(\rootname-2-2.north)+(0.1,0)$) {$leaf_5$};

                \probist{\rootname-2-1-1}{0.7}{0.3}
                \node[anchor=south] at ($(\rootname-2-1-1.north)+(-0.1,0)$) {$leaf_3$};
                \probist{\rootname-2-1-2}{0.2}{0.8}
                \node[anchor=south] at ($(\rootname-2-1-2.north)+(0.1,0)$) {$leaf_4$};

                \node[below=0.6em of \rootname-1-1] (label-1) {
                    \textcolor{classtruecolor}{0.598}/\textcolor{classfalsecolor}{0.032}
                };
                \node[below=0.6em of \rootname-1-2] {
                    \textcolor{classtruecolor}{0.00}/\textcolor{classfalsecolor}{0.07}
                };

                \node[below=0.6em of \rootname-2-1-1] {
                    \textcolor{classtruecolor}{0.019}/\textcolor{classfalsecolor}{0.008}
                };

                \node[below=0.6em of \rootname-2-1-2] (h) {/};
                \node[right=0.7em of h.west, anchor=east] (trueprob) {\textcolor{classtruecolor}{0.001}};
                \node[right=-0.7em of h.east, anchor=west] (falseprob) {\textcolor{classfalsecolor}{0.002}};

                \node[below=0.6em of \rootname-2-2] (h) {/};
                \node[right= 0.7em of h.west, anchor=east] {\textcolor{classtruecolor}{0.003}};
                \node[right=-0.7em of h.east, anchor=west] {\textcolor{classfalsecolor}{0.267}};

            \end{pgfonlayer}

            \begin{pgfonlayer}{background}
                \begin{scope}[->]
                    \draw (\rootname) edge (\rootname-1);
                    \drawbetween{\rootname}{\rootname-1}{0.7}{0.5}

                    \draw (\rootname-1) edge[midway, left]  (\rootname-1-1);
                    \drawbetween{\rootname-1}{\rootname-1-1}{0.9}{0.5}

                    \draw (\rootname-1) edge (\rootname-1-2);
                    \drawbetween{\rootname-1}{\rootname-1-2}{0.1}{0.5}

                    \draw (\rootname) edge (\rootname-2);
                    \drawbetween{\rootname}{\rootname-2}{0.3}{0.5}

                    \draw (\rootname-2) edge (\rootname-2-1);
                    \drawbetween{\rootname-2}{\rootname-2-1}{0.1}{0.5}
                    
                    \draw (\rootname-2) edge (\rootname-2-2);
                    \drawbetween{\rootname-2}{\rootname-2-2}{0.9}{0.5}

                    \draw (\rootname-2-1) edge (\rootname-2-1-1);
                    \drawbetween{\rootname-2-1}{\rootname-2-1-1}{0.9}{0.3}

                    \draw (\rootname-2-1) edge (\rootname-2-1-2);
                    \drawbetween{\rootname-2-1}{\rootname-2-1-2}{0.1}{0.3}

                    \node[fit= (\rootname) (\rootname-1-1) (\rootname-2)] {};

                    \coordinate (help) at (\rootname |- \rootname-2-1-1.south);
                    \node at ($(help) + (0,-2.0)$) {$
                        \begin{aligned}
                            \mathunderline{classtruecolor}{P(pos)} &=
                            \mathunderline{classtruecolor}{0.598 + 0.019 + 0.001 + 0.003} = \mathunderline{classtruecolor}{0.621} = 1 - \mathunderline{classfalsecolor}{0.379}\\
                                                                   &= 1 - \mathunderline{classfalsecolor}{(0.032 + 0.07 + 0.008 + 0.002 + 0.267)} = 1 - \mathunderline{classfalsecolor}{P(neg)}
                        \end{aligned}
                $};
                \end{scope}
            \end{pgfonlayer}
        \end{scope}
}

\begin{scope}

    \begin{pgfonlayer}{foreground}
        \node[anchor=north] at ($(row-0 -| column-1) + (-4.4,0)$) (rules-header) {\textbf{Rules}};
        \node[anchor=north] (rules) at ($(rules-header.south) + (0,0.3)$) {
            \begin{minipage}{0.6\textwidth}
                \begin{align*}
& 0.95 :: d_1.\quad  pos \impliedby d_1 \land leaf_1 .\quad
neg \impliedby \neg d_1 \land leaf_1 .\\
& \phantom{neg \impliedby \neg burglary} \hdots\\
&0.01 :: d_5 \quad pos \impliedby d_5 \land leaf_5.\quad
neg \impliedby \neg d_5 \land leaf_5.\\[0.7em]
&leaf_1 \impliedby \phantom{\neg} burglary \land alarm .\\
& \phantom{neg \impliedby \neg burglary} \hdots \\
&leaf_5 \impliedby \neg burglary \land \neg earthquake .\\
                \end{align*}
            \end{minipage}
            };
    \end{pgfonlayer}

{
        \begin{pgfonlayer}{foreground}

            \node[anchor=north] (header) at ($(row-0 -| column-1) + (4.4,0)$) {$\mathbf{\mathcal{PF}}$};
            \node[anchor=north] (example-1) at ($(header.south) + (0,0.3)$) {
                \begin{minipage}{0.35\textwidth}
                \begin{align*}
&0.7 :: burglary.\\
&0.1 :: earthquake.\\
&0.9 :: alarm.\\
                \end{align*}
                \end{minipage}
            };

        \end{pgfonlayer}
}
\end{scope}

\end{tikzpicture}
  }
  \caption{\label{fig:probDT} 
    At the bottom we have a graphical representation of a probabilistic decision tree.
    Note that in this example, both $alarm$ literals, although in different branches, refer to the same test.
    Furthermore, we assign probability distributions over the classes $pos$ and $neg$ to each leaf node (cf. Equations \ref{eq:lambdatwo} and \ref{eq:lambdathree}).
    At the top we see the \ac{PDT} expressed as a probabilistic logic program. We provide a translation algorithm in \autoref{ssec:nsdtc} for the general neurosymbolic case.
  }
\end{figure}
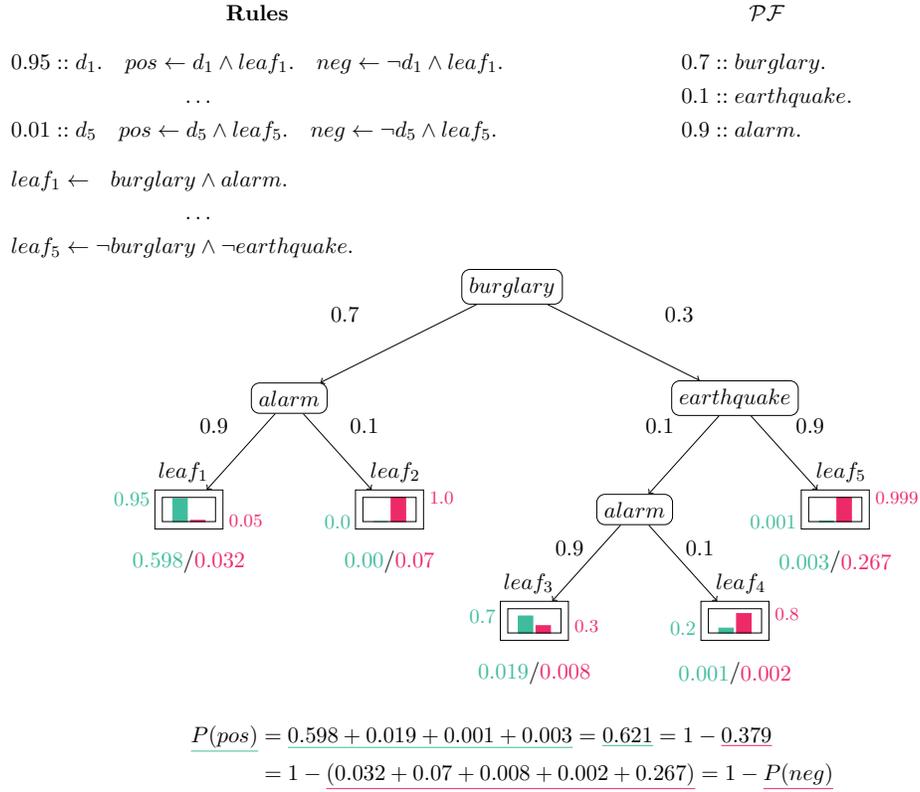

So far we have assumed that all tests in $\cal F$ are deterministically true or false.
We now lift this restriction by adopting probabilistic logic instead and use the concept of a probabilistic fact~\cite{fierensInferenceLearningProbabilistic2015, deraedtProbLogProbabilisticProlog2007}. Probabilistic facts generalize (non-probabilistic) facts in that they allow atoms to be true with a probability. In a logic program we write such probabilistic facts as annotated facts $p_i\probcoco f_i$ stating that $f_i$ is true with probability $p_i$. Consequently, a (non-probabilistic) fact is a probabilistic fact annotated with probability $1.0$.
To incorporate this idea in a \ac{BDT}, we render all tests (i.e., internal nodes) probabilistic.
We exemplify this in ~\autoref{fig:probDT}.

Similar to \acp{BDT}, a probabilistic Boolean Decision Tree can be evaluated with probabilistic logic. Formally, a probabilistic logic program consists of a set of rules $R$ and a set of probabilistic facts $\mathcal{PF}$ of the form $p_i \probcoco f_i$.
Probabilistic logic programs induce a probability distribution on the set of interpretations of the logic theory, \ie{} the set of its possible worlds, cf. \cite{deraedtProbLogProbabilisticProlog2007}.
Probabilistic logic programs can be viewed as expressive probabilistic graphical models. 
Indeed, the probabilistic facts are marginally independent just like the nodes without parents in a Bayesian network, and dependencies are introduced through the deterministic rules~\cite{de2015probabilistic}. 
For our purposes, it suffices to define the probability $P(q)$ of an atom $q$ in a logic program $L = R \cup \mathcal{PF}$ consisting of rules $R$ and probabilistic facts $\mathcal{PF} = \{p_1 \probcoco f_1, \dots , p_n::f_n\}$. 

We define the probability of a subset $M \subseteq \mathcal{F}=\{f_1, ... , f_n\}$ containing probabilistic facts as
\begin{align}
  \label{formula:weightedModels}
  P(M) &= \prod_{f_i \in M} p_i \prod_{f_i \in \mathcal{PF} \backslash M} (1- p_i),
  \end{align}
which we then use to define the probability of a query $q$
  \begin{align}
  P(q) &= \sum_{M \cup L \models q} P(M).
  \label{formula:weightedModels2}
\end{align}
Computing the probability in Eq. \ref{formula:weightedModels2} is a well-studied problem known as \ac{WMC}~\cite{darwicheKnowledgeCompilationMap2002}. Probabilistic logic programming languages such as ProbLog~\cite{deraedtProbLogProbabilisticProlog2007, fierensInferenceLearningProbabilistic2015} address this problem through knowledge compilation where the logic is compiled into so-called arithmetic circuits \cite{darwicheKnowledgeCompilationMap2002}, \ie{} hierarchically structured product and sum operations) that allow for inference in polytime in size of the circuit.
Although \ac{WMC} is in general a computationally hard problem (\#P-complete), the probability of a query to a \ac{PDT} can be efficiently computed (in the absence of background knowledge), in linear time in the size of the decision tree. The branches in the decision tree are by construction mutually exclusive.  This makes that the sums are naturally disjoint and the notorious disjoint sum problem does not arise.

The probability of an arbitrary example falling into a specific leaf equals the probability of the body of the corresponding rule being true.
For instance, consider the decision tree in \autoref{fig:probDT}. The probability of falling into the third leaf in the logic program for the example $e_1$ is computed as:
\begin{align*}
  & \phantom{{}={}} P(\neg burglary \land earthquake \land alarm)\\
  &=0.3 \cdot 0.1 \cdot 0.9\\
  &=0.027
\end{align*}
with the set of probabilistic facts $\mathcal{PF}$ shown in \autoref{fig:probDT}. This is remeniscent of Quinlan's probabilistic decision trees \cite{quinlan1990probabilistic}.

By using probabilistic tests instead of deterministic ones on the internal nodes of the \ac{BDT}, a single example no longer falls deterministically into a specific leaf, but probabilistically into all leaves. It is then useful to make the class assignments in the leaves probabilistic themselves.
We accomplish this by replacing the deterministic class assignments $ pos \impliedby leaf_i$ in the \ac{BDT} with probabilistic rules
\begin{align}
     \delta_i::d_i.
          \label{eq:lambdaone}\\
 pos \leftarrow d_i \wedge leaf_i.
  \label{eq:lambdatwo}
\\
 neg \leftarrow \neg d_i \wedge leaf_i.
\label{eq:lambdathree}
\end{align}

\noindent where $\delta_i \in [0,1]$. This statement means that the probability of $pos$ is $\delta_i$ and  probability of $neg$ is $1- \delta_i$ when $leaf_i$ evaluates to true.

We estimate $\delta_i$ for a leaf $leaf_i$ from a set of labeled data $\mathcal{E}$ using the relative frequencies of the examples falling into leaf, as is standard in Decision Tree induction.
However, in our probabilistic tree, each example falls only fractionally in the leaf $leaf_i$ with probability $P(leaf_i |e)$. This explains why we estimate $\delta_i$ as:
\begin{align}
    \delta_{i} = P(pos \mid leaf_i) &= \cfrac{P(pos , leaf_i)}{P(leaf_i)}
    \nonumber
    \\ 
    & = \cfrac{
    \sum_{e} P(pos , leaf_i \mid e) \cdot P(e)
    }{
    \sum_{e} P(leaf_i \mid e) \cdot P(e)
    }
    \nonumber
    \\
    &= \cfrac{
    \sum_e P(pos \mid leaf_i ) \cdot P(leaf_i \mid e) \cdot  P(e)}{\sum_e P(leaf_i \mid e) \cdot P(e)}
    \nonumber
    \\
    &= \cfrac{\mathbb{E}_e \bigl[P(pos \mid leaf_i ) \cdot P(leaf_i \mid e) \bigr]}{
    \mathbb{E}_e \bigl[P(leaf_i \mid e)\bigr]}
    \nonumber
    \\
    &\approx
     \cfrac{
        \sum_{e \in \mathcal{E}} P(pos \mid leaf_i ) \cdot P(leaf_i \mid e)
    }
    {
        \sum_{e \in \mathcal{E}} P(leaf_i \mid e)
    }
    \label{eq:lambda_approx}
\end{align}
where $\sum_e$ denotes the sum over all possible examples and $\sum_{e \in \mathcal{E}}$ restricts the sum to those in  the dataset $\mathcal{E}$. We compute the probabilities by querying a logic program consisting of the rules $R$ encoding the decision tree and its leaves, and the probabilistic facts corresponding to the example $e$.

\newcommand{\earthquake}{\raisebox{-3px}{\includegraphics[height=1.5\fontcharht\font`\B]{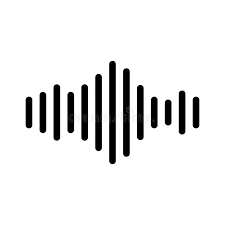}}}
\newcommand{\window}{\raisebox{-2px}{\includegraphics[height=1.5\fontcharht\font`\B]{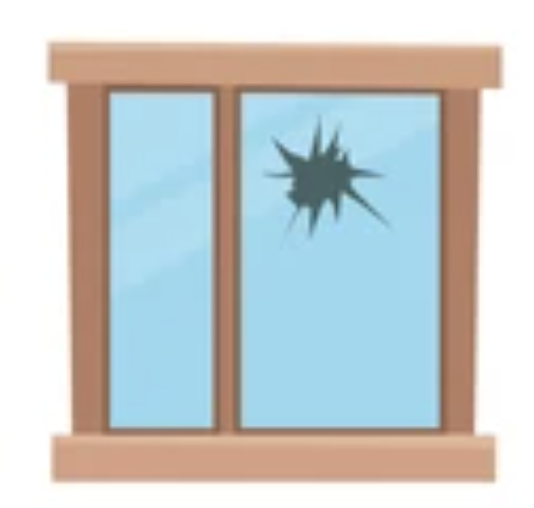}}}
\newcommand{\Conj}{\kappa}

\section{Neurosymbolic Decision Trees}
\label{sec:NLDT}

\subsection{Probabilistic Neural Decision Trees}
\label{ssec:neuralfact}

We now present a neural extension that allows for the probabilities of specific facts to depend on a specific example. 
The key idea is simple, and is based on the neural extension introduced in DeepProbLog \cite{manhaeveNeuralProbabilisticLogic2021,manhaeveDeepProbLogNeuralProbabilistic2018}:
instead of using unconditional probabilities on the probabilistic facts, we now use conditional probabilities parametrized using neural nets.
More formally, a (probabilistic) neural fact is an expression of the form $nn(\mathbf{x}):: f$,
where $nn$ is a neural network receiving input $\mathbf{x}$. It returns the probability that $f$ holds given the input $\mathbf{x}$, \ie{} $P(f \mid \mathbf{x})$. Neural facts, when evaluated, are interpreted as conditional probabilistic facts.
As such, we can also interpret probabilistic facts as neural facts with an empty conditioning set. Thereby neural facts strictly generalize probabilistic facts (and deterministic ones).

As an example, consider again example $e_1$ and the \ac{PDT} introduced in \autoref{fig:probDT}. However, we now replace the probabilistic facts $burglary$ and $earthquake$ with neural ones whose probabilities are predicted by the two \acp{NN} $nn_{b}$ and $nn_{eq}$, respectively. In return, we obtain a neurosymbolic program with one probabilistic fact and two neural facts, which we group together in the set 
\begin{align*}
    \mathcal{NF}_{e1}&= \{ 0.9 :: alarm, ~ nn_{b}(\window) :: burglary,   ~ nn_{eq}(\earthquake) :: earthquake  \}
\end{align*}
with $\window$ and $\earthquake$ being subsymbolic features for example $e1$ and
\begin{align*}
 nn_{b}(\window)&= P(burglary|\window)  
 \\
nn_{eq}(\earthquake)&= P(earthquake | \earthquake).
\end{align*}
Note how we index the set of neural facts with the subscript $e_1$ as every example potentially induces a different (conditional) probability for the neural facts.

\subsection{Neurosymbolic Background Knowledge}
\label{ssec:neuralrule}

One distinguishing feature of \ac{NeSy} is its ability to incorporate background knowledge. 
We allow for this in \acp{NLDT} by extending the set of possible tests with neurosymbolic rules. 
For ease of exposition, we introduce neurosymbolic background knowledge by means of an example.

Consider a classifier that is given two images of playing cards as input and has to decide which of the two cards has the higher rank. 
The task then boils down to correctly classifying each card and comparing their ranks. We can readily encode this problem using the concept of a neural predicate~\cite{manhaeveDeepProbLogNeuralProbabilistic2018}.
\begin{align*}
	&nn(Image, Rank, [1,\ldots, 10])::rank(Image,Class).
    \\
    &smaller(Image1, Image2)  \leftarrow \\
    & \qquad  \qquad rank(Image1, Rank1), rank(Image2,Rank2), Rank1 < Rank2.
\end{align*}
In the first line we have a neural predicate, which encodes a categorical distribution whose events are integer-valued and take values in the $1$ to $10$ range. The neural predicate hence encodes the probability distribution $P(Rank | Image)$ with $Rank \in [1,\cdots, 10]$ (assuming we limit ourselves to numeral cards).
We then use the neural predicate in the body of the rule in the second line, where we let the neural predicate predict the ranks of the two cards and symbolically compare them. The head of the rule $smaller(Image1, Image2)$ is then true if the comparison in the body evaluates to true. As the predictions of the neural predicate are probabilistic, the comparison of the two predicted ranks is probabilistic, which means that the rule is probabilistically true or false as well.
We can then use these probabilistic rules (in combination with neural predicates) as tests in neurosymbolic decision trees. We refer the reader to \cite{manhaeveDeepProbLogNeuralProbabilistic2018,manhaeveNeuralProbabilisticLogic2021} for a detailed account of neural rules and predicates, and their use in background knowledge.

\subsection{Neurosymbolic Decision Trees as Classifiers}
\label{ssec:nsdtc}

We define an \ac{NLDT} as a binary tree whose internal nodes consist of tests and whose leaves encode conditional probability distributions $P(c \mid leaf_i)$, with $c\in \{pos, neg \}$ and $leaf_i$ denoting the $i$-th leaf. A test belongs to either of the following four kinds \begin{mylist}
    \item a determinsitic fact (\autoref{ssec:BCDT}),
    \item a probabilitistic fact (\autoref{ssec:PDT}),
    \item a neural fact (\autoref{ssec:neuralfact}),
    \item or a neural rule (\autoref{ssec:neuralrule}).
\end{mylist}
We can use \acp{NLDT} as neurosymbolic classifiers by asking for the probability
\begin{align*}
     P(c \mid e) = \sum_i P(c \mid leaf_i) \cdot P(leaf_i \mid e).
\end{align*}

Instead of writing a custom inference algorithm to compute these probabilities,  we can translate \acp{NLDT} to DeepProbLog programs, and leverage its inference mechanism. This translation will also be valuable when learning the parameters of the neural tests (neural facts and rules) as we perform this again using DeepProbLog's inference engine as an oracle.
The translation of an \ac{NLDT} to a DeepProbLog program works as follows.\\\\
Initialize the set $L$ that represents the logic program with $L=\mathcal{T}$. \\
Then, for each leaf $leaf_i$
\begin{enumerate}[itemsep=0pt, parsep=0pt, topsep=0pt]
    \item determine the conjunction of tests $\test_1 \wedge ... \wedge \test_n$ forming the path from the root to the respective leaf $leaf_i$ and add a rule $leaf_i \leftarrow \test_{i1} \wedge ... \wedge \test_{in}$ to the set $L$.
    \item add the probabilistic fact $\delta_i :: d_i$ to the set $L$.
    \item add the rule $ pos \leftarrow d_i \land leaf_i.$ to the set $L$.
    \item add the rule $ neg \leftarrow \neg d_i \land leaf_i.$ to the set $L$.
\end{enumerate}

\section{Learning Neurosymbolic Decision Trees}
\label{sec:LearningNDTs}

Assume that we are given a set of examples each labeled with either $pos$ or $neg$. The question is then whether we can learn an \ac{NLDT} given a predefined set of tests $        \mathcal{T}$.
As we would like to learn a neurosymbolic model the examples consist of Boolean and subsymbolic features (\eg{} images).
In the context of neurosymbolic logic programming, Boolean features correspond to logical facts and subsymbolic features can be used in the conditioning sets of neural facts and neural predicates.

Learning an \ac{NLDT} that classifies examples then consists of finding an adequate tree structure and learning the parameters in the neural tests (neural facts and neurosymbolic rules).
This stands in contrast to many other \ac{NeSy} system, \eg{} DeepProbLog~\cite{manhaeveDeepProbLogNeuralProbabilistic2018} and NeurASP~\cite{yangNeurASPEmbracingNeural2020}, where only the parameters are being learned.

\subsection{\namealgo{} -- the Learning Algorithm}
\label{ssec:NDTLearner}

The underlying principle of \namealgo{} resembles that of vanilla top-down induction tree algorithms for \acp{DT} like ID3~\cite{quinlanInductionDecisionTrees1986} and its first order extension TILDE~\cite{blockeelTopinductionFirstorderLogical1998}.
The idea is to refine a classifier by recursively replacing leaf nodes with fresh tests until a prespecified stopping criterion is reached. We give the pseudocode for \namealgo{} in \autoref{alg:inducingTree}.

If $\mathcal{E}$ and $\mathcal{T}$ are the set of (labeled) examples and possible tests, we invoke the learner with $\textbf{\namealgo{}}(\mathcal{E}, \mathcal{T}, \Conj = true)$. In each recursion, new tests are appended to $\Conj$ representing the root-leaf path to a new node $N$ which we initialize as a leaf node in the Lines \ref{alg:newNode}-\ref{alg:estimateProb}.
A leaf node stores $\Conj$ and a $\delta$ estimated by our function \textbf{estimate\_probability} whose implementation follows our \autoref{eq:lambda_approx}. The new node is returned as a leaf node and the algorithm terminates when it meets a prespecified stopping criterion (Line \ref{algo:stopCrit}), \ie{} a maximum tree depth or when extending the node would not improve the scoring function, see below for details.

In case the stopping criterion is not met, we continue by refining $N$ into an internal node which requires assigning a test to $N$. For that reason, \namealgo{} trains all neural tests in $\mathcal{T}$ (Line \ref{algo:train_test}). Then the best among the trained ones is selected (Line \ref{algo:chooseBest}) and removed from $\mathcal{T}$. Note that \textbf{train\_tests} only affects neural and probabilistic tests. We explain the functions \textbf{train\_tests} and \textbf{best\_test} in further detail in \autoref{sec:learning} and \autoref{sec:scoring}. 

Next, we add two new child nodes $N.left$ and $N.right$ to $N$ and finalize $N$'s refinement into an internal node. For that reason, we compute a new conjunction $\Conj_l$ (Line \ref{algo:conj_l}) for $N.left$ and calculate an $\mathcal{E}_l \subseteq{} \mathcal{E}$ by removing every example $e \in \mathcal{E}$ with $P(\Conj_l \mid e) < \epsilon$ (Line \ref{algo:filterOne}) with $\epsilon$ being a pre-defined hyperparameter. We compute $\Conj_r$ and $\mathcal{E}_r$ similarly for $N.right$. We use the new example sets and conjunctions to extend both child nodes by calling \namealgo{} on both nodes (Line \ref{algo:extendLeft}-\ref{algo:extendRight}). Subsequentualy, we return $N$.

\begin{algorithm}
	\SetAlgoNoEnd{}
	\DontPrintSemicolon{}
	\SetKwProg{Fn}{Function}{:}{}
	\Fn{$\textbf{\namealgo{}}( \mathcal{E}: \text{Examples},\ \mathcal{T}: \text{Tests}, \ \kappa : \text{Conjunction}$) }{
        $N := \textbf{new\_node}()$\label{alg:newNode}
        \\
        $N.\Conj := \Conj$
        \\
        $N.\delta$ := \textbf{estimate\_probability}($\mathcal{E}, N.\Conj$)\label{alg:estimateProb} 
            \\
		\If{ stopping criterion is met\label{algo:stopCrit}}{
            \Return $N$
		}
		\Else(){

            $\mathcal{T} := \textbf{train\_tests}(\mathcal{E},\ \mathcal{T}, \ N)$
            \label{algo:train_test}
            \\
            $N.\test := \textbf{best\_test} (\mathcal{E},\ \ \mathcal{T},\ N )$
            \label{algo:chooseBest}
            \\
            $\mathcal{T} := \mathcal{T} \setminus \{ N.\test \}$
            \\
            $\Conj_l := N.\Conj \land N.\test$ \label{algo:conj_l}
            \\
            $\Conj_r := N.\Conj \land \neg N.\test$ \label{algo:conj_r}
            \\
			 $\mathcal{E}_l := \Bigl\{ e \in \mathcal{E}
				\mid
			P \left(\Conj_l \mid e \right) \geq\epsilon \Bigr\}$\label{algo:filterOne} \;

			$\mathcal{E}_r := \Bigl\{ e \in \mathcal{E}
				\mid
				P \left(\Conj_r \mid e \right) \geq\epsilon \Bigr\}$ \label{algo:filterTwo}\;
                
			$N.left := \textbf{\namealgo{}}
            \left(\mathcal{E}_l,\ \mathcal{T},\ \Conj_l \right) $
            \label{algo:extendLeft}
            \\
			$N.right := \textbf{\namealgo{}}
            \left(\mathcal{E}_r,\ \mathcal{T},\ \Conj_r \right) $\label{algo:extendRight}
            \\
            \Return $N$
		}
	}
\vspace{\baselineskip}
\caption{\namealgo{}}
\label{alg:inducingTree}
\end{algorithm}

\subsection{Loss Function}
\label{sec:learning}

Every time we add a new test to an \ac{NLDT} using \autoref{alg:inducingTree} we train all the neural tests in $\mathcal{T}$. Following the approach of DeepProbLog~\cite{manhaeveDeepProbLogNeuralProbabilistic2018}, we would use the standard \ac{CE} loss to perform the optimization for a candidate test $\test\in\mathcal{T}$.
\begin{align*}
    CE(N,\test) &=  \sum_{e\in \mathcal{E}_N} 
    \underbrace{\Bigl[
    log\bigl(P\left(\test \mid N.\Conj,  e \right) \cdot \mathbbm{1}_{l_e=pos}
    \Bigr]}_{ce_{pos}(N,e, \test)} +
    \\
     & \qquad \sum_{e\in \mathcal{E}_N}
    \underbrace{\Bigl[
	log\bigl(P\left(  \neg \test \mid N.\Conj, e \right) \cdot \mathbbm{1}_{l_e=neg} \Bigr]}_{ce_{neg}(N,e, \test)}
    \\
    &=  \sum_{e\in \mathcal{E}_N}
    \left[ce_{pos}(N,e, \test) + ce_{neg}(N,e, \test)\right],
\end{align*}
where we compute $P\left(N.\Conj \land \test \mid  e \right)$ and $P\left(N.\Conj \land \neg \test \mid  e \right)$ using oracle calls to DeepProbLog, and where $\mathbbm{1}_{l_e=pos}$ indicates whether the label of an example is $pos$ or not (analogous for $\mathbbm{1}_{l_e=neg}$). We also explicitly indicate with a subscript on $\mathcal{E}_N$ that we only take into consideration the subset of examples that fall into node $N$ after applying the filtering steps in Lines \ref{algo:filterOne} and \ref{algo:filterTwo} in \autoref{alg:inducingTree}.

However, in the setting of training (neural) tests in an \ac{NLDT} this overlooks two issues. First, examples only fall fraction-wise into the lower nodes due to probabilistic splitting. Second, with an increasing depth class imbalances arise because the classification accuracy is improving.
As a consequence, the deeper a test is situated in the tree the more imbalanced the dataset becomes.
We remedy these issues by replacing the standard cross-entropy as used in DeepProbLog with a weighted cross-entropy loss $\widetilde{CE}(N)$ when training a test in node $N$. 
\begin{align*}
    \widetilde{CE}(N, \test) = 
\sum_{e\in \mathcal{E}_N}
    \left[w_{pos}(N,e) \cdot ce_{pos}(N,e,\test) + w_{neg}(N,e) \cdot ce_{neg}(N,e,\test)\right],
\end{align*}
where the weights $w_{c}(N,e)$ ($c\in \{pos , neg\}$) are defined as follows:
\begin{align*}
    w_c(N,e)
    &=
     P(N.\Conj \mid e) \cdot \frac{1}{\underset{=2}{\underbrace{|\{pos, neg\}|}} \cdot P(c \mid N.\Conj) }
    \\
     &=
     P(N.\Conj \mid e) \cdot \frac{1}{2 \cdot (N.\delta \cdot \mathbbm{1}_{c=pos} + (1 - N.\delta) \cdot \mathbbm{1}_{c=neg}) }
\end{align*}
with  $P( N.\Conj \mid e)$ being the probability of node $N$ being reached given example $e$ -- thereby handling the issue of fractional examples,
and where the factor
$1/( 2 \cdot P(c \mid N.\Conj))$ addresses the issue of a class imbalance of two classes. The computation of the class weights follows the implementation of scikit-learn\cite{scikit-learn} and is inspired by \cite{kingLogisticRegressionRare2001}.

\subsection{Scoring function}
\label{sec:scoring}

Once the neural tests have been learned we need to pick one test to add to the \ac{NLDT}. To this end, we compute for each test (including non-neural ones) a score. The test with the highest score is then selected and added to the \ac{NLDT}.
In order to select tests to be added, deterministic decision trees commonly use the \ac{IG} of a test $\test$ in node $N$~\cite{cover1999elements}:
\begin{align*}
    IG_N(\test)
    &=
    H(Class \mid N.\Conj)
    \\
    & \qquad
    -
    \underset{\approx P(\test \mid N.\kappa)}{\underbrace{\cfrac{|\mathcal{E}_{N}(\test)|}{ |\mathcal{E}_{N}| }}}
    H(Class \mid N.\Conj \land \test)
    -
    \underset{\approx P(\neg \test \mid N.\kappa)}{\underbrace{\cfrac{|\mathcal{E}_{N}(\neg \test)|}{ |\mathcal{E}_{N}| }}}
    H(Class \mid  N.\Conj \land \neg \test),
\end{align*}
where $H(Class \mid \cdot )$
denotes the conditional entropy of the random variable $Class$ and is computed as follows:
\begin{align*}
    H( Class  \mid \phi)
    &
    =
    \sum_{c\in \{pos, neg \}} P(Class=c\mid \phi) \log_2 P(Class=c\mid \phi)
    \\
    &=
    \delta_\phi \log_2 \delta_\phi
    +
    (1-\delta_\phi) \log_2 (1-\delta_\phi),
\end{align*}
with $\phi$ denoting a propositional formula and which we approximate using \autoref{eq:lambda_approx}.
Furthermore, the ratios $\nicefrac{|\mathcal{E}_{N}(\test)|}{ |\mathcal{E}_{N}| }$ and $\nicefrac{|\mathcal{E}_{N}(\neg \test)|}{ |\mathcal{E}_{N}| }$ serve as approximations for $P(\test \mid N.\Conj)$ and $P(\neg \test \mid N.\Conj)$ where:  
\begin{align*}
    \mathcal{E}_N(\phi) = \{ e \in \mathcal{E}_N \mid N.\kappa \land \phi \models e \}.
\end{align*}

For probabilistic (neural) tests, we replace the relative example counts with probabilities of the examples falling into the respective branches: 
\begin{align*}
    \widetilde{IG}_N(\test)
    &=
    H(Class \mid N.\Conj)
    \\
    & \quad
    -
    \cfrac{S_{N}(\test)}{ S_{N} }
    H(Class \mid N.\Conj \land \test)
    -
    \cfrac{S_{N}(\neg \test)}{ S_{N} }
    H(Class \mid  N.\Conj \land \neg \test),
\end{align*}
with $S_N$ and $S_N(\cdot)$ as
\begin{align*}
    S_{N} &=
    \sum_{e \in \mathcal{E}_N} P\Bigl( N.\Conj \mid e \Bigr)\\
	S_{N}(\phi)
    &=
    \sum_{e  \in \mathcal{E}_N}
    P\Bigl( N.\Conj \wedge \phi  \mid e \Bigr) \qquad (\phi \in \{\test, \neg \test \}).
\end{align*}
Our estimations $P(\test \mid N.\Conj, e) \approx \nicefrac{S_{N}(\test)}{S_N}$ and $P(\neg \test \mid N.\Conj, e) \approx \nicefrac{S_{N}(\test)}{S_N}$ can be derived following \autoref{eq:lambda_approx}.
Finally, we define \textbf{best\_test} for picking the best test in node~$N$ (c.f. Line \ref{algo:chooseBest} in \autoref{alg:inducingTree}) as
\begin{align*}
    {\mathbf{best\_test}} (\mathcal{E}, \mathcal{T}, N)
    =
    \argmax_{\test \in \mathcal{T}} \widetilde{IG}_N(\test).
\end{align*}
Note that for the sake of notational clarity, we drop the explicit dependency of $\widetilde{IG}_N(\test)$ on $\mathcal{E}$.

\section{Related Work}
\label{sec:related_work}

As \acp{NLDT} belong on the one hand to the family of decision trees and on the other hand to the set of \ac{NeSy} techniques, we separate the related work into two subsections.  We first characterize \acp{NLDT} and \namealgo{} in the context of decision trees (\autoref{sec:related_work_dt}) and secondly in the \ac{NeSy} context (\autoref{sec:related_work_nesy}).

\subsection{Decision Trees}
\label{sec:related_work_dt}

{
  \newcommand{\skipdiff}{\hskip 1.5cm}
\begin{table}
    \centering
    \caption{Comparison to related works implementing \acp{DT}.}
    \label{tab:relatedWork}
    \resizebox{\textwidth}{!}{
      \begin{tabular}{lc@{\skipdiff}cc@{\skipdiff}c@{\skipdiff}c@{\skipdiff}c}
        \toprule
         & Searches & \multicolumn{2}{c@{\skipdiff}}{Tests} & Learned & Probabilistic & Background\\
         & Input Features  & symbolic & subsymbolic & Structure & Splitting & Knowledge\\
         \midrule
        \multicolumn{7}{c}{Soft Decision Trees} \\
        \midrule
        \citet{tannoAdaptiveNeuralTrees2019}  & \unchecked & \unchecked & \checked & \checked & \checked  & \unchecked \\
        \citet{irsoySoftDecisionTrees2012}  & \unchecked & \unchecked & \checked & \checked & \checked & \unchecked  \\
        \citet{frosstDistillingNeuralNetwork2017a}  & \unchecked & \unchecked & \checked & \unchecked & \checked & \unchecked  \\
        \citet{wanNBDTNeuralBackedDecision2021}  & \unchecked & \unchecked & \checked & \checked & \checked & \unchecked  \\
        \citet{luoSDTRSoftDecision2021}  & \unchecked & \unchecked & \checked & \unchecked & \checked  & \unchecked \\

        \midrule
        \multicolumn{6}{c}{\ac{DT} learning through parameter learning} \\
        \midrule

        \citet{martonGradTreeLearningAxisAligned2024a}  & \checked & \checked & \unchecked & \unchecked & \unchecked & \unchecked  \\
        \citet{yangDeepNeuralDecision2018}              & \unchecked & \checked & \unchecked & \unchecked & \unchecked & \unchecked  \\
        \midrule
         \acp{NLDT}  & \checked & \checked & \checked & \checked & \checked & \checked \\
      \end{tabular}
    }
\end{table}
}

We identify five main dimensions along which the different approaches to specifying and learning decision trees can be characterized which we also reflect in our \autoref{tab:relatedWork}.
\begin{enumerate}[itemsep=0pt, parsep=0pt]

    \item Whether a relevant subset of the features is selected by the learner (like in decision trees), or whether the learner always uses the full set of features. While the second option avoids the need to find the best features, it also decreases the interpretability of the model and may require more data. 
    \item
Whether the tests can be symbolic and/or subsymbolic. In the latter case, the test is carried out as (part of) a neural network.
\item Whether the structure tree is pre-fixed and provided as input to the learner, or whether the structure itself is learned.  By pre-fixing the tree, one only has to optimize the parameters over the whole tree and effectively avoids problems like \textit{weaker splits hiding stronger splits}~\cite{bertsimasOptimalClassificationTrees2017}. On the other hand, those approaches usually also fix aspects related to selecting features at each node, the order of selecting tests, and determining the depth of a tree.
\item 
Whether the splits can be used in a probabilistic manner. Probabilistic approaches compute the probability that an example falls into a leaf, and allow to aggregate the probabilities over several leaves to compute the class membership, although some approaches \cite{frosstDistillingNeuralNetwork2017a, tannoAdaptiveNeuralTrees2019} select the maximum probability path. Deterministic approaches can only use one single path for predicting.
\item Whether background knowledge in the form of rules can be used to define and derive further features from the inputs for use by the learner. 
\end{enumerate}

We observe in \autoref{tab:relatedWork} that related work can be distinguished into roughly two categories. The first can be called soft decision trees, which are essentially a type of neural net that is structured as a decision tree and where the tests are softened. This is related to the probabilistic splitting that we discussed.
The second category learns (deterministic) \acp{DT} by reformulating the search over symbolic splits and input features as a parameter learning problem.  

From this analysis, it follows that \acp{NLDT} are uniquely positioned in the context of recent related work on decision trees through their distinct neurosymbolic approach.

\subsection{Neurosymbolic Learning}
\label{sec:related_work_nesy}

We briefly discuss how our work relates to other \ac{NeSy} structure learning approaches. Our discussion is led by the work of \citet{marraStatisticalRelationalNeurosymbolic2024} separating \ac{NeSy} learning works into four classes:
\begin{mylist}
\item neurally guided structure search,
\item program sketching,
\item structure learning via parameters,
\item and learning structure implicitly
\end{mylist}. Interestingly, \acp{NLDT} together with \namealgo{} do not fit in any of these categories and, instead, are a new group that alternates between learning structure and parameters. 
We now discuss the differences between NDTs and these classes of related work.

\paragraph{Neurally guided structure search} This class describes approaches~\cite{luNeurallyGuidedStructureInference2019, ellisLearningLibrariesSubroutines2018} where structure learning is guided by a function approximator that replaces the typical uninformed search usually found in structure learning. For instance, \acp{NN} can be used to score parts in the search space and accelerate the structure learning, or they can even generate symbolic representations that are added to the theory. These approaches usually require large amounts of data \cite{marraStatisticalRelationalNeurosymbolic2024, ellisWriteExecuteAssess2019}. In contrast, \acp{NLDT} are \ac{NeSy} programs with learned parameters and structure but still rely on an exhaustive and uninformed search.  Furthermore, neurally guided search has mainly been used to find symbolic structures, and to the best of the authors' knowledge, not yet to learn neurosymbolic models themselves. 

\paragraph{Program sketching} These approaches~\cite{manhaeveNeuralProbabilisticLogic2021} usually fix parts the program and keep a few parts open that still need to be learned or filled in. This requires the user to already have an understanding of the target program and only learns the parts of the structure. In contrast, \acp{NLDT} are fully learned from scratch.  

\paragraph{Structure learning via parameters} Here, works~\cite{manhaeveDeepProbLogNeuralProbabilistic2018, evansLearningExplanatoryRules2018, siSynthesizingDatalogPrograms2019} reformulate the structure learning task into an optimization problem solved by parameter learning. 
One example is to start from a set of clauses and learning weights for each of them. While learning the weights (parameters), clauses can become irrelevant (and get low weights), but they will never be removed from the theory which results in noisy theories. In contrast, \acp{NLDT} are induced bottom-up and result in clear theories.

\paragraph{Learning structure implicitly} These approaches~\cite{marraNeuralMarkovLogic2021} do not learn a human-readable structure of a logic program but instead an implicit neural net modeling relations between data. In contrast, \acp{NLDT} learn human-readable structures.

\newcommand{\UCIPure}{UCI\textsubscript{pure}}
\newcommand{\UCIonehot}{UCI\textsubscript{one-hot}}
\newcommand{\UCIMNIST}{UCI\textsubscript{MNIST}}

\newcommand{\Concept}{\ensuremath{\mathcal{C}}\xspace}
\newcommand{\concept}{\ensuremath{c}\xspace}

\newcommand{\neuralPredicate}[1]{\ensuremath{\mathcal{T}^{nf}_{#1}}}

\newcommand{\optimalPool}[1]{\ensuremath{\mathcal{T}^{opt}_{#1}}\xspace}

\newcommand{\generalPool}{\ensuremath{\mathcal{T}^{opt}_{\cup \concept}}\xspace}

\newcommand{\badPool}[1]{\ensuremath{\mathcal{T}^{bad}_{#1}\xspace}}

\newcommand{\EQOne}{\textbf{How do \acp{NLDT} compare to \acp{NN}?}}
\newcommand{\EQTwo}{\textbf{How does including tests that implement background knowledge affect the induction of \acp{NLDT}?}}
\newcommand{\EQThree}{\textbf{How can we reuse trained neural tests to induce more complex rules?}}

\newcommand{\ImSuit}[1]{ImSuit#1}
\newcommand{\ImRank}[1]{ImRank#1}
\newcommand{\SymSuit}[1]{Suit#1}
\newcommand{\Bool}[1]{Bool}
\newcommand{\SymRank}[1]{Rank#1}

\section{Experimental Evaluations}
\label{sec:experiments}

\namealgo{} induces \acp{NLDT} given symbolic and subsymbolic features. We are interested in answering the following experimental questions.
\begin{enumerate}[parsep=0pt, itemsep=0pt]
	\item \EQOne
	\item \EQTwo
	\item \EQThree
\end{enumerate}

\subsection{Datasets}

We introduce two benchmarks for our evaluation. In the first, we conduct a binary classification task on tabular datasets consisting of binary features, which are either represented by symbols (1 and 0) or subsymbols (%
\includegraphics[height=\fontcharht\font`\B]{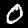}
and 
\includegraphics[height=\fontcharht\font`\B]{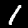}). The second benchmark involves learning hidden concepts from playing cards. 

\paragraph{UCI Dataset}

In the first benchmark, we evaluate our \acp{NLDT} on synthetic versions of UCI classification datasets. For clarity, we will call the unprocessed UCI datasets \UCIPure{}. We discuss briefly how we generated symbolic and subsymbolic versions of datasets in \UCIPure{}.

We generated binary classification datasets of binary features from \UCIPure{} that we name \UCIonehot{}. For simplicity, we assigned dataset-wise a $d \in [2,4]$ and created for each non-binary feature $d$ new binary features. The numerical features were transformed by entropy-based binning~\cite{fayyadMultiIntervalDiscretizationContinuousValued1993} and one-hot encoding, while the categorical features were transformed into $d$ binary features by randomly assigning the unique values in the categorical attribute to one of the newly created binary features. As for conversion of multi-class to binary classification datasets, we chose the most-frequent class as positive and all other classes as negatives.

Our \UCIMNIST{} datasets are variations of the \UCIonehot{} datasets where all binary features are replaced by MNIST~\cite{lecunGradientbasedLearningApplied1998} images of 1
(\eg{} \includegraphics[height=\fontcharht\font`\B]{./graphics/experiments/experiment_1/mnist/1/0.png}, \includegraphics[height=\fontcharht\font`\B]{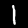})
and 0
(\eg{} \includegraphics[height=\fontcharht\font`\B]{./graphics/experiments/experiment_1/mnist/0/0.png}, \includegraphics[height=\fontcharht\font`\B]{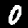})
respectively. We used the MNIST train dataset to generate training data and the MNIST test dataset for generating our test sets.

\paragraph{Card Sets}
Our second benchmark involves recognizing whether a sequence of two cards fulfills a hidden concept, cf. \citet{dietterichApplyingGeneralInduction1980}. 

First, we generated two sets of card representations as $C = \{(s_i, r_i)\}_{1 \leq i \leq 20000}$ with $s_i \in S=\{diamond, clubs, hearts, spades\}$ and $r_i \in [1, 8]$\footnote{We also experimented with $r_i=7$, $r_i=9$ and $r_i=10$ and observed qualitatively identical results.} where the suits and ranks are evenly distributed. By choosing 8 ranks, we simulate a card setting similar to a Piquet pack that contains four face and four non-face cards. The first set of cards is used to create the training and validation set, while the second set constitutes the test set.

We generated subsymbolic versions of our symbolic card representations by substituting all suit symbols with EMNIST~\cite{cohenEMNISTExtendingMNIST2017} letter images (A-D), and the ranks with MNIST~\cite{lecunGradientbasedLearningApplied1998} images. We used images of the MNIST and EMNIST test sets to create the cards for testing and their training sets are used for creating cards for training and validating.
Thus, these two sets of cards do not share images of cards. Subsequently, we color our cards either white or red depending on the suit. We draw 10\% of cards from our training card set to generate a validation card set.

\paragraph{Eleusis Dataset}

For our second benchmark, we chose the game Eleusis as an environment. Eleusis is a card game often associated with inductive reasoning and has been used to evaluate symbolic systems~\cite{dietterichInductiveLearningStructural1981, dietterichApplyingGeneralInduction1980, henleChangingRulesLegacy2021}. In this game, players lay down sequences of cards that satisfy an unknown rule. The players only receive feedback on whether a sequence of cards is valid or not, and they have to induce the hidden concept based on the previous turns. For the sake of clarity, we call these rules \textit{hidden concepts}.
We limit this experiment to learning hidden concepts from windows of two cards.

We use our train, validation, and test card sets to generate datasets representing hidden concepts introduced by \citet{dietterichApplyingGeneralInduction1980}. We use most hidden concepts except for the ones that are not expressible in propositional logic but require first-order logic. However, we also added two more complex hidden concepts which are disjunctions of hidden concepts of the work of \cite{dietterichApplyingGeneralInduction1980}. The complete list of hidden concepts can be found in \ref{tab:eleusisRules}. An example can be seen in Table \autoref{fig:experimentEleusisExample}. We generate different versions for each hidden concept by using 7 different random seeds.

For each combination of hidden concept and seed, we generate a train-, test- and validation dataset where each example consists of 2 cards and a label that represents whether the concept was fulfilled. The train and test dataset consists of 1000 samples, while the validation dataset consists of 200 samples.

\subsection{Experiment: Comparison with \acp{NN}}

In this experiment, we address our first experimental question: \textbf{\EQOne}  on our UCI and Eleusis datasets.

\subsubsection{UCI}

We use \UCIonehot{} and \UCIMNIST{} as dataset environments.
The \UCIonehot{} serves as a baseline to \UCIMNIST{} and indicates how much training neural predicates affects the performance.

As for the \ac{NN}'s architecture, we oriented ourselves on a survey related to tabular data conducted by \citeauthor{borisovDeepNeuralNetworks2022}~\cite{borisovDeepNeuralNetworks2022}. Here, a simple \ac{MLP} outperformed most other neural non-hybrid models on categorical data, while being easier to train and adapt. We use the hyperparameters provided by the authors.\footnote{https://github.com/kathrinse/TabSurvey, accessed 17.04.24} For symbolic input, we use the original architecture as used in the survey. For subsymbolic input, we replace each input neuron with an MNIST classifier of an output size of 2. We train our models over 50 epochs.

As for our \acp{NLDT}, we adapt our set of tests based on whether it learns on \UCIonehot{} or \UCIMNIST{}. When we induce on \UCIonehot, the tests are simple atoms encoding whether a single symbolic features is true. When we induce on \UCIMNIST{}, we train neural predicates for each subsymbolic feature. The neural predicates use MNIST networks~\cite{manhaeveDeepProbLogNeuralProbabilistic2018} in our neural predicates which we train over 20 epochs each.

We perform 10-fold cross-validation on the \UCIonehot{} and the \UCIMNIST{} dataset. We measure the accuracy of the model and default accuracy on each split. The default accuracy is the accuracy that a model would reach by only predicting the most-frequent class in the training dataset. 
\\[1em]

\paragraph{Results}

We present our results in \autoref{tab:experiment1Results}. The \ac{MLP} and the \ac{NLDT} show an accuracy close to the default accuracy on the last five presented datasets. This indicates that no model could learn on any of these datasets indicating that our one-hot encoding obtained by pre-processing was not sufficient to learn a meaningful model. On the remaining datasets, our \acp{NLDT} outperform the \ac{MLP} on \UCIMNIST{} by 0.34 on average in accuracy. However, the \ac{MLP} outperforms the \ac{NLDT} on the \UCIonehot{} dataset by 0.02 points on average in accuracy.

The difference in the performance loss between the \UCIonehot{} and \UCIMNIST{} datasets differs between the models. The \ac{NLDT}'s accuracy decreases by 0.02, while the \ac{MLP}'s decrease in accuracy is about 0.40 on average. Overall, the \ac{NN}'s accuracy is relatively close to the default accuracy indicating a failure in learning.

\begin{table*}[h!]
	\caption{\label{tab:experiment1Results}The table shows the results of our first
		experiment on \UCIMNIST{} and \UCIonehot{}. We measure performance by accuracy and provide the results over all 10-folds through mean and standard deviation ($\pm$). Furthermore, we provide the accuracies for our models with respect to their input.
		The order of the rows indicates the general learning success of the data.}
	\resizebox{\textwidth}{!}{

\ifstandalone
\usepackage{booktabs}
\usepackage{graphicx}

\begin{document}
\fi

\begin{tabular}{lccc||c|cccc}
\toprule
 & \multicolumn{3}{c}{Property} & Default & \multicolumn{2}{c}{MLP} & \multicolumn{2}{c}{Tree} \\
 & \#attributes & \parbox{5em}{\#test} & \parbox{5em}{\#train} & Accuracy & \parbox{5em}{\UCIonehot} & \parbox{7em}{\UCIMNIST} & \parbox{5em}{\UCIonehot} & \parbox{7em}{\UCIMNIST} \\
\midrule
breast-cancer & 43 & 68 & 614 & 0.65$\pm$0.01 & 0.97$\pm$0.02 & 0.57$\pm$0.05 & 0.95$\pm$0.02 & 0.94$\pm$0.03 \\
car-evaluation & 21 & 172 & 1555 & 0.70$\pm$0.00 & 0.99$\pm$0.00 & 0.57$\pm$0.05 & 0.86$\pm$0.01 & 0.74$\pm$0.02 \\
congressional & 16 & 23 & 208 & 0.53$\pm$0.03 & 0.95$\pm$0.04 & 0.55$\pm$0.09 & 0.97$\pm$0.03 & 0.96$\pm$0.04 \\
credit-approval & 15 & 65 & 587 & 0.55$\pm$0.01 & 0.84$\pm$0.03 & 0.43$\pm$0.05 & 0.86$\pm$0.03 & 0.86$\pm$0.03 \\
dermatology & 34 & 35 & 322 & 0.69$\pm$0.01 & 0.97$\pm$0.02 & 0.57$\pm$0.12 & 0.91$\pm$0.05 & 0.87$\pm$0.04 \\
diabetes & 16 & 52 & 468 & 0.62$\pm$0.00 & 0.96$\pm$0.03 & 0.61$\pm$0.04 & 0.90$\pm$0.05 & 0.87$\pm$0.03 \\
iris & 12 & 15 & 135 & 0.67$\pm$0.00 & 1.00$\pm$0.00 & 0.67$\pm$0.00 & 1.00$\pm$0.00 & 1.00$\pm$0.00 \\
mushroom & 22 & 564 & 5079 & 0.62$\pm$0.00 & 1.00$\pm$0.00 & 0.51$\pm$0.03 & 0.99$\pm$0.00 & 0.94$\pm$0.02 \\
wine & 13 & 17 & 160 & 0.60$\pm$0.02 & 0.98$\pm$0.03 & 0.58$\pm$0.11 & 0.92$\pm$0.05 & 0.90$\pm$0.05 \\
zoo & 16 & 10 & 90 & 0.59$\pm$0.03 & 0.98$\pm$0.04 & 0.61$\pm$0.08 & 1.00$\pm$0.00 & 1.00$\pm$0.00 \\
\midrule
blood-transfusion & 16 & 74 & 673 & \underline{0.76$\pm$0.00} & 0.79$\pm$0.03 & 0.76$\pm$0.00 & 0.76$\pm$0.00 & 0.76$\pm$0.01 \\
contraceptive & 27 & 147 & 1325 & \underline{0.57$\pm$0.00} & 0.69$\pm$0.04 & 0.46$\pm$0.03 & 0.64$\pm$0.01 & 0.57$\pm$0.04 \\
hepatitis & 51 & 8 & 72 & \underline{0.84$\pm$0.06} & 0.84$\pm$0.06 & 0.84$\pm$0.06 & 0.86$\pm$0.10 & 0.84$\pm$0.06 \\
statlog-german & 64 & 100 & 900 & \underline{0.70$\pm$0.00} & 0.72$\pm$0.03 & 0.58$\pm$0.09 & 0.70$\pm$0.00 & 0.70$\pm$0.00 \\
yeast & 32 & 148 & 1335 & \underline{0.69$\pm$0.00} & 0.70$\pm$0.04 & 0.61$\pm$0.05 & 0.69$\pm$0.00 & 0.68$\pm$0.01 \\
\bottomrule
\end{tabular}

\ifstandalone
\end{document}
\fi

	}
\end{table*}

\subsubsection{Eleusis}

For the \ac{NN} we chose an encoder-decoder architecture where the decoder predicts two classes (pos and neg) on the output of four separate encoders. In turn, each encoder predicts on one of four images representing our two cards (2 suits and 2 ranks). An encoder is a three-layered \ac{MLP} on top of the latent space of a MNIST network~\cite{manhaeveDeepProbLogNeuralProbabilistic2018} and decreases layer-wise the output size. We implement four encoders to motivate the \ac{NN} to learn different features for every image (rank or suit of one card).
We train each \ac{NN} on $50$ epochs with Adam as optimizier and a learning rate of $1e^{-3}$.

We induce each \ac{NLDT} from a set of neural facts where each fact depends on either the suits or ranks of both cards. The \ac{NN} used in our neural atoms is a simple encoder-decoder architecture following the idea of our \acp{MLP} explained in the previous paragraph. We train all our neural tests for 20 epochs, use Adam as optimizier with a learning rate of $1e^{-3}$.
\\[1em]

\paragraph{Results}

\begin{table}
	\caption{\label{tab:experiment1Eleusis} The table shows the results on the Eleusis Datasets. We present our results through the mean to standard deviation ($\pm$) of the F1 scores over all seeds and number of ranks for each hidden concept.}
	\resizebox{\textwidth}{!}{
		\begin{tabular}{l|ll|ll}
\toprule
 & \multicolumn{2}{c}{Class Positive} & \multicolumn{2}{c}{Class Negative} \\
 & neural\_basline(44M) & \ac{NLDT} & neural\_basline(44M) & \ac{NLDT} \\
\midrule
hidden\_modulo\_simple & $0.729\pm0.102$ & $\mathbf{0.876\pm0.014}$ & $0.213\pm0.106$ & $\mathbf{0.752\pm0.015}$ \\
hidden\_order\_simple & $0.476\pm0.078$ & $\mathbf{0.695\pm0.035}$ & $0.761\pm0.039$ & $\mathbf{0.850\pm0.019}$ \\
color\_parity\_rule & $0.641\pm0.063$ & $\mathbf{0.977\pm0.007}$ & $0.485\pm0.055$ & $\mathbf{0.976\pm0.007}$ \\
increase\_suits & $0.826\pm0.031$ & $\mathbf{0.973\pm0.027}$ & $0.125\pm0.096$ & $\mathbf{0.917\pm0.090}$ \\
suit\_order & $0.898\pm0.025$ & $\mathbf{0.985\pm0.004}$ & $0.803\pm0.038$ & $\mathbf{0.976\pm0.008}$ \\
rank\_order & $0.637\pm0.056$ & $\mathbf{0.877\pm0.015}$ & $0.551\pm0.062$ & $\mathbf{0.844\pm0.020}$ \\
alternating\_parity & $0.479\pm0.167$ & $\mathbf{0.871\pm0.057}$ & $0.497\pm0.102$ & $\mathbf{0.883\pm0.045}$ \\
alternating\_faces & $0.510\pm0.070$ & $\mathbf{0.740\pm0.194}$ & $0.443\pm0.185$ & $\mathbf{0.764\pm0.177}$ \\
\bottomrule
\end{tabular}

	}
\end{table}

We present the results in \autoref{tab:experiment1Eleusis}. The \ac{NLDT} outperforms the \ac{NN} on average in every hidden concept. We find the means of our F1 scores of our \acp{NLDT} on each hidden concept to be higher and observe the standard deviation to be lower on almost all hidden concepts.

\subsubsection{Discussion}

We can now answer our experimental question: \textbf{\EQOne}. {\it  In our experiments \acp{NLDT} outperform the \acp{NN} and the difference in performance increases when Boolean attributes are replaced by images. } This is especially obvious on the \UCIonehot{} and \UCIMNIST{} datasets. The drop in the \acp{NLDT}' performance from \UCIonehot{} to \UCIMNIST{} is relatively low compared with the neural counterpart.
A possible explanation is that our top-down induction of \acp{NLDT} searches for subsets of features and learns to process them separately, while a \ac{NN} has to learn simultaneously on all features.

Next, we discuss the stability in inducing \acp{NLDT} on different data distributions. In our UCI experiment, we observe that the standard deviation of our \ac{NLDT} is quite similar between \UCIonehot{} and \UCIMNIST{}. Arguably, replacing Boolean values with simple images has little effect on the induction of \acp{NLDT}. This becomes even more obvious when we consider that this holds on the dermatology dataset with $322$ training samples and 34 different attributes and supports the stability of inducing our trees. Our Eleusis experiment supports this observation where most standard deviations are below $5\%$.

With regard to training times, the \ac{NN} clearly outperforms the \ac{NLDT} when trained on the CPU of the same machine\footnote{tested on a a \textit{Intel(R) Xeon(R) E-2176G CPU @ 3.70GHz}}. For instance, training an \ac{MLP} on the wine and dermatology datasets (from \UCIMNIST{}) requires 56 and 150 seconds respectively, whereas training an \ac{NLDT} takes 23 and 118 minutes. In the Eleusis dataset, training an \ac{MLP} on a CPU takes around 5 minutes on all hidden concepts, while the induction of a \ac{NLDT} takes between 1.5 to 2.2 hours.
In part, this discrepancy can be explained through the differences between structure learning and parameter learning. For \acp{NN} only the parameters need to be estimated, while for \acp{NLDT} its structure also has to be learned.
We also note that \acp{NN} benefit from highly optimized implementations that have emerged during the last decade.
Attaining similar gains in \ac{NeSy} AI is currently an active research field, but the first results seem very promising \citep{maene2025klay}. Ultimately, \namealgo{} will also benefit from this trend.


\begin{figure}
	\resizebox{\textwidth}{!}{
		\input{./graphics/experiments/experiment_eleusis/example.tikz}
	}
	\caption{\label{fig:experimentEleusisExample} The figure shows the components of our experiment. Each Eleusis dataset contains examples of four images representing two subsequent cards. The labels present whether they fulfill a hidden concept or not. The set of tests contains DeepProbLog rules that include untrained neural predicates like $suit$ or $rank$.}
\end{figure}

\subsection{Experiment: Background Knowledge}

In all our prior experiments, \namealgo{} only induced \acp{NLDT} from sets of tests containing only neural or symbolic facts, and, therefore, we have not considered neural rules. We address this issue with our experimental question: \EQTwo.

We chose Eleusis as the setting for our experiment and learn each hidden concept from the previous experiment by inducing \acp{NLDT} from four different classes of $\mathcal{T}$:

\begin{enumerate}
\setlength{\itemsep}{0pt}
\item Our first class \neuralPredicate{} only uses two neural facts that depend on either the suits or the ranks of a card as tests:
\begin{align*}
    nn_r(\ImRank{1}, \ImRank{2}) &:: rel\_rank(\ImRank{1}, \ImRank{2}).\\
    nn_s(\ImSuit{1}, \ImSuit{2}) &:: rel\_suit(\ImSuit{1}, \ImSuit{2}).
\end{align*}
where each probability is computed by a \ac{NN} that receives both images as input.
We provide a further illustration in \autoref{fig:eleusisNeuralAtoms}. \neuralPredicate{} does not depend on the hidden concept that shall be learned.
\item The class \optimalPool{\concept} assigns for each concept $\concept$ a set of tests which we designed with the intent that \namealgo{} only has to learn how to combine the tests and to train them to learn $\concept$. For that purpose, we formulate \ac{NeSy} rules that can express the hidden concept on a symbolic level. For instance, assume the hidden concept alternate\_faces:
\begin{align*}
faces(Card_0) \not = faces(Card_1).
\end{align*}
that states that two subsequent cards' ranks need to alternate between face cards and non-face cards. To learn this concept, we only define one \ac{NeSy} rule as test.
\begin{align*}
alternate\_attr&(\ImRank{1}, \ImRank{2}) \leftarrow\\
	    &\phantom{\leftarrow}\quad rank\_attribute(\ImRank{1}, Attr1) ,\\ &\phantom{\leftarrow}\quad rank\_attribute(\ImRank{2}, Attr2) ,\\
							  &\phantom{\leftarrow}\quad Attr1 \not= Attr2 
\end{align*}
where $rank\_attribute$ is a neural predicate that has to be learned that maps an image (\ie{} $\ImRank{1}$) to an Boolean attribute (\ie{} $Attr2$). The set of tests will be:
\begin{align*}
    \optimalPool{alternate\_faces} = \{ alternate\_attr \}    
\end{align*}
Here, \namealgo{} has only one test to choose from which simplifies the task to learning the neural predicate $alternate\_attr$.
Another example can be seen in \autoref{fig:experimentEleusisExample} where the hidden concept is a conjunction of two tests that are in the set of tests.
\item \generalPool{} constructs the set of tests as$$
\generalPool{} =
\bigcup_{\concept \in \Concept}
\optimalPool{\concept}
$$ with $\Concept$ being the set of hidden concepts. \generalPool{} contains all untrained tests defined in our class \optimalPool{} and is the most exhaustive set in this experiment. The challenge for \namealgo{} is to detect which untrained tests are useful to learn the hidden concept.
\item For each hidden concept $\concept$, we define \badPool{\concept}  as:
\begin{align*}
\badPool{\concept} =
 \Bigl(
    \generalPool{} \cup \neuralPredicate{}
\Bigr)
\setminus
\optimalPool{\concept}.
\end{align*}
More intuitively, we start from \generalPool{} and remove the tests that we designed to learn the specific hidden concept $\concept$. We intend the remaining tests to either be neural rules that are \textit{bad} for learning $\concept$ or neural facts. This set of tests  is used to assess whether \namealgo{} would avoid \textit{bad} neural rules and compensate them with neural facts. 
\end{enumerate}

\paragraph{Results}

\begin{table}
    \caption{\label{tab:eleusisAggResults}This table shows the ratio of our models to our optimal baseline \optimalPool{\concept}. Lower values indicate that the models on average receive lower F1 scores while higher values indicate a better performance.}
	\resizebox{\textwidth}{!}{

\ifstandalone
  \usepackage{tikz, pgf}
  \usepackage{booktabs}

  \standaloneenv{new_image}

  \begin{document}
\fi

\begin{tabular}{l|lll|lll}
  \toprule
 & \multicolumn{3}{c}{True} & \multicolumn{3}{c}{False} \\
 & NLDT(\badPool{\concept}) & NLDT(\neuralPredicate{}) & NLDT(\generalPool{}) & NLDT(\badPool{\concept}) & NLDT(\neuralPredicate{}) & NLDT(\generalPool{}) \\
 \midrule
alternating\_faces & $0.76 \pm 0.20$ & $0.78 \pm 0.19$ & $0.99 \pm 0.03$ & $0.80 \pm 0.17$ & $0.80 \pm 0.17$ & $0.99 \pm 0.02$ \\
alternating\_parity & $0.88 \pm 0.03$ & $0.89 \pm 0.06$ & $0.98 \pm 0.02$ & $0.89 \pm 0.03$ & $0.90 \pm 0.05$ & $0.98 \pm 0.02$ \\
color\_parity\_rule & $0.99 \pm 0.01$ & $0.99 \pm 0.01$ & $1.00 \pm 0.00$ & $0.98 \pm 0.01$ & $0.99 \pm 0.01$ & $1.00 \pm 0.00$ \\
hidden\_modulo\_simple & $0.92 \pm 0.08$ & $0.97 \pm 0.02$ & $0.97 \pm 0.02$ & $0.89 \pm 0.16$ & $0.97 \pm 0.04$ & $0.98 \pm 0.03$ \\
hidden\_order\_simple & $0.94 \pm 0.07$ & $0.93 \pm 0.07$ & $0.95 \pm 0.06$ & $0.99 \pm 0.04$ & $0.99 \pm 0.04$ & $0.99 \pm 0.03$ \\
increase\_suits & $0.98 \pm 0.03$ & $0.98 \pm 0.03$ & $1.00 \pm 0.00$ & $0.92 \pm 0.09$ & $0.94 \pm 0.09$ & $1.00 \pm 0.01$ \\
rank\_order & $0.88 \pm 0.01$ & $0.90 \pm 0.02$ & $0.99 \pm 0.03$ & $0.86 \pm 0.01$ & $0.88 \pm 0.02$ & $0.98 \pm 0.03$ \\
suit\_order & $0.99 \pm 0.00$ & $0.99 \pm 0.01$ & $1.00 \pm 0.01$ & $0.98 \pm 0.01$ & $0.99 \pm 0.01$ & $1.00 \pm 0.01$ \\
\midrule
average & $0.92\pm0.10$ & $0.93\pm0.10$ & $0.98\pm0.03$ & $0.91\pm0.11$ & $0.93\pm0.09$ & $0.99\pm0.02$ \\
\end{tabular}

\ifstandalone
  \end{document}
\fi

}
\end{table}

\begin{table}
    \caption{This table reports the aggregated class-wise F1 scores over all experiments with different seed. We report the aggregated results by the mean to standard deviation}
    \label{tab:experimentEleusisAllComp}
    \centering
    \resizebox{\textwidth}{!}{

       \begin{tabular}{l|llll}
\toprule
concept \concept & NDT(\badPool{\concept}) & NDT(\neuralPredicate{}) & NDT(\generalPool{}) & NDT(\optimalPool{\concept}) \\
\midrule
\multicolumn{5}{c}{F1 Score (Positives)}\\
\midrule
hidden\_modulo\_simple%
 & $0.851\pm0.076$ & $0.881\pm0.022$ & $0.880\pm0.018$ & $\mathbf{0.916}\pm0.012$ \\
hidden\_order\_simple%
 & $0.680\pm0.022$ & $0.700\pm0.029$ & $0.694\pm0.020$ & $\mathbf{0.707}\pm0.029$ \\
color\_parity\_rule%
 & $0.945\pm0.009$ & $0.956\pm0.009$ & $0.961\pm0.013$ & $\mathbf{0.969}\pm0.007$ \\
increase\_suits%
 & $0.953\pm0.074$ & $0.978\pm0.009$ & $0.987\pm0.004$ & $\mathbf{0.990}\pm0.003$ \\
suit\_order%
 & $0.983\pm0.004$ & $0.985\pm0.004$ & $0.991\pm0.005$ & $\mathbf{0.993}\pm0.003$ \\
rank\_order%
 & $0.830\pm0.016$ & $0.845\pm0.012$ & $0.886\pm0.061$ & $\mathbf{0.938}\pm0.012$ \\
alternating\_parity%
 & $0.715\pm0.080$ & $0.702\pm0.092$ & $\mathbf{0.946}\pm0.010$ & $0.942\pm0.012$ \\
alternating\_faces%
 & $0.733\pm0.119$ & $0.716\pm0.123$ & $0.933\pm0.014$ & $\mathbf{0.935}\pm0.021$ \\
\midrule
\multicolumn{5}{c}{F1 Score (Negatives)}\\
\midrule
hidden\_modulo\_simple%
 & $0.690\pm0.157$ & $0.742\pm0.069$ & $0.754\pm0.037$ & $\mathbf{0.794}\pm0.037$ \\
hidden\_order\_simple%
 & $0.848\pm0.012$ & $\mathbf{0.858}\pm0.012$ & $\mathbf{0.858}\pm0.014$ & $0.849\pm0.008$ \\
color\_parity\_rule%
 & $0.944\pm0.007$ & $0.955\pm0.008$ & $0.960\pm0.012$ & $\mathbf{0.968}\pm0.007$ \\
increase\_suits%
 & $0.838\pm0.277$ & $0.933\pm0.031$ & $0.960\pm0.015$ & $\mathbf{0.969}\pm0.009$ \\
suit\_order%
 & $0.971\pm0.009$ & $0.975\pm0.008$ & $0.985\pm0.008$ & $\mathbf{0.989}\pm0.005$ \\
rank\_order%
 & $0.805\pm0.015$ & $0.821\pm0.014$ & $0.867\pm0.070$ & $\mathbf{0.927}\pm0.013$ \\
alternating\_parity%
 & $0.734\pm0.066$ & $0.730\pm0.091$ & $\mathbf{0.947}\pm0.009$ & $0.943\pm0.010$ \\
alternating\_faces%
 & $0.725\pm0.128$ & $0.696\pm0.157$ & $\mathbf{0.933}\pm0.016$ & $\mathbf{0.933}\pm0.022$ \\
\bottomrule
\end{tabular}

    }
\end{table}

We report our aggregated results as the mean to standard deviation of all F1 scores of all seeds in \autoref{tab:experimentEleusisAllComp}. Further, we report an aggregation statistics in \autoref{tab:eleusisAggResults} displaying the ratios of F1 scores between our trained models to our optimal set of tests \optimalPool{}.

\autoref{tab:experimentEleusisAllComp} shows that \optimalPool{} achieves on average the best F1 scores with the lowest standard deviation followed by \generalPool{}. \autoref{tab:eleusisAggResults} reveals that both settings are close in terms of performance. In contrast, \badPool{} receives the worst performance on almost all hidden concepts but is often close to \neuralPredicate{}. Especially noticable, is how bad \neuralPredicate{} and \badPool{} perform on alternating\_parity, rank\_order and alternating\_faces compared to the other two sets.

We further investigate which neural tests were used in \acp{NLDT} induced from \badPool{}. We found that from all tests that were used in \acp{NLDT} only four were \ac{NeSy} rules. Three of those four \ac{NeSy} rules were used in the second layer of the tree and only had 50-100 examples to train. Further, we found that two of those were used in the same tree.

\paragraph{Discussion}
W.r.t. our second experimental question: \textbf{\EQTwo}, 
we observe that \optimalPool{} - as the bool that contains background knowledge designed for each rule - receives the best results. From our experiments, we conclude that
\textit{providing appropriate background knowledge for inducing a concept leads to a better induction of better \acp{NLDT}.}\\
\noindent The positive effect of providing useful background knowledge is also underlined by the fact that the results on \generalPool{} are similar to \optimalPool{} but outperform \neuralPredicate{}.

Another aspect that we identified is that \acp{NLDT} appear to be reliable in not using tests whose background knowledge is counter-productive for learning a hidden concept. In \badPool{}, \namealgo{} chose most of the time a neural atom and avoided other tests. Consequently, our tree learner appears to be successful in recognizing when a new relationship has to be learned from scratch, which is not formulated in our background knowledge, yet.

\subsection{Experiment: Reuse Neural Tests}

In this experiment, we address our third experimental question: \EQThree{}. For that purpose, we use our insights from the previous experiment. Inducing \acp{NLDT} for the hidden concept hidden\_order\_simple has shown relatively poor results. However, this hidden concept is a disjunction of the two hidden concepts suit\_order and rank\_order.

In this experiment, we reuse the neural tests received from suit\_order and rank\_order to induce \acp{NLDT} where we \begin{mylist}
    \item use no trained neural tests,
    \item reuse one neural test received from training suit\_order or rank\_order,
    \item or use both trained neural tests.
\end{mylist} We induce each \ac{NLDT} with the same seven seeds as in the previous experiment.

\paragraph{Results}

Our results are shown in \autoref{fig:eleusisAblation}. 
The results show that including one neural test leads to better results than not using one at all. Using both trained tests leads to the best results. Further, we observe differences in performance depending on which trained neural test we use. Reusing the neural test from rank\_order leads to better results and shows similar results as when we reuse both neural tests.

\paragraph{Discussion}

We now discuss our experimental question: \EQThree{}.
The results show that adding trained neural tests can be beneficial for inducing a new hidden concept. Especially noticable is the effect of adding the neural test received from rank\_order which receives similar results to using both trained neural tests. This indicates that learning the order of suits is relatively simple compared to learning the order of the ranks. This can be explained by the aspect that our \acp{NLDT} tend to first use the trained neural test and train the untrained ones in lower layers. As discussed in \autoref{sec:LearningNDTs}, we have less data samples in nodes in lower layers. Further, learning suit\_order means learning an order on 4 elements while learning the ranks\_order requires learning an order on 8 elements.  

\begin{table}
    \caption{The table displays the results of learning hidden\_order\_simple by using suit\_order and rank\_order as trained or untrained neural tests. We report the class-wise F1 scores over all seeds as mean to standard deviation.}
    \label{fig:eleusisAblation}
    \centering

    \begin{tikzpicture}
  \pgfdeclarelayer{foreground}
  \pgfdeclarelayer{background}
  \pgfsetlayers{background, foreground}

  \begin{pgfonlayer}{foreground}

    \matrix[matrix of nodes, nodes in empty cells,
    row sep=1.4em, column sep=1em
    ]%
    (table) {
              & untrained & trained\\
    untrained & $0.863\pm0.024$ & $0.915\pm0.018$ \\
    trained & $0.978\pm0.002$ & $0.981\pm0.004$ \\
    };

    \node[anchor=south] (suit-order) at ($(table-1-2)!.5!(table-1-3) + (0,0.3)$) {suit\_order};
    \node[anchor=south, rotate=90, transform shape] (rank-order) at ($(table-2-1.west)!.5!(table-3-1.west) + (-0.3,0)$) {rank\_order};

  \end{pgfonlayer}

  \begin{pgfonlayer}{background}
  \begin{scope}
    \coordinate (north-max) at (suit-order.north);
    \coordinate (south-max) at (rank-order.west); 
    \coordinate (east-max) at (table.east); 
    \coordinate (west-max) at (rank-order.north); 

    \coordinate (help) at ($(table-2-1.east)!0.5!(table-2-2.west)$);
    \draw[draw=black] (north-max -| help) -- (south-max -| help);

    \coordinate (help) at ($(table-1-1.south)!0.5!(table-2-1.north)$);
    \draw[draw=black] (help -| east-max) -- (help -| west-max);

  \end{scope}
  \end{pgfonlayer}

\end{tikzpicture}
    
\end{table}

\subsection{Summary of the experiments}
Our experiments demonstrated that top-down induced \acp{NLDT} outperform \acp{MLP} on tabular datasets with image features. The \ac{MLP} showed little to no learning success, while \namealgo{} could successfully induce and train \acp{NLDT}. Arguably, the \ac{MLP} struggled to detect relevant features, while the weak supervision used to induce \acp{NLDT} was sufficient to detect which features are of interest. Furthermore, we demonstrated in our experiments the advantages of including background knowledge in the tests and highlight the advantage of \acp{NLDT} being modular systems where parts can be reused, for instance, in a continual learning setting.

\section{Conclusion}
\label{sec:conclusion}

We introduced \acp{NLDT} as neurosymbolic soft decision trees and proposed \namealgo{} as a
natural extension and upgrade of the popular decision tree learning paradigm. The neurosymbolic component is based on the DeepProbLog \cite{manhaeveNeuralProbabilisticLogic2021} family of systems. 
\namealgo{} is a neurosymbolic learning system that cannot only learn the parameters of neural networks but also the symbolic structure of such models from scratch.  At the same time, \namealgo{} is able to deal with both symbolic and subsymbolic features, and it can use background knowledge and weak supervision in the induction process. The experiments show that \namealgo{} performs better than purely neural approaches to tackle the same tasks.  

Decision tree learning, thanks to its simplicity, has proven to be very effective, efficient and popular. Numerous extensions of decision trees exist and it would be interesting to exploit such extensions also for neurosymbolic decision trees.

\section*{Acknowledgements}
This research has received funding from  
the Wallenberg AI, Autonomous Systems and Software Program (WASP) funded by the Knut and Alice Wallenberg Foundation,
and from the European Research Council (ERC) under the European Union's Horizon Europe research and innovation programme (grant agreement DeepLog n°101142702).

The computations and data handling for this work were enabled by the Tetralith resource at the
National Supercomputer Centre, provided by the Knut and Alice Wallenberg Foundation, under
projects “NAISS 2024/22-1221”.
We also want to thank the Ericsson Research Data Center (ERDC) for sharing their resources and providing their support.

\bibliographystyle{plainnat}

\bibliography{main}

\begin{thebibliography}{40}
\providecommand{\natexlab}[1]{#1}
\providecommand{\url}[1]{\texttt{#1}}
\expandafter\ifx\csname urlstyle\endcsname\relax
  \providecommand{\doi}[1]{doi: #1}\else
  \providecommand{\doi}{doi: \begingroup \urlstyle{rm}\Url}\fi

\bibitem[Bertsimas and Dunn(2017)]{bertsimasOptimalClassificationTrees2017}
Dimitris Bertsimas and Jack Dunn.
\newblock Optimal classification trees.
\newblock \emph{Machine Learning}, 106\penalty0 (7):\penalty0 1039--1082, July
  2017.
\newblock ISSN 0885-6125, 1573-0565.
\newblock \doi{10.1007/s10994-017-5633-9}.

\bibitem[Blockeel and
  De~Raedt(1998)]{blockeelTopinductionFirstorderLogical1998}
Hendrik Blockeel and Luc De~Raedt.
\newblock Top-down induction of first-order logical decision trees.
\newblock \emph{Artificial Intelligence}, 101\penalty0 (1):\penalty0 285--297,
  May 1998.
\newblock ISSN 0004-3702.
\newblock \doi{10.1016/S0004-3702(98)00034-4}.

\bibitem[Borisov et~al.(2022)Borisov, Leemann, Se{\ss}ler, Haug, Pawelczyk, and
  Kasneci]{borisovDeepNeuralNetworks2022}
Vadim Borisov, Tobias Leemann, Kathrin Se{\ss}ler, Johannes Haug, Martin
  Pawelczyk, and Gjergji Kasneci.
\newblock Deep {{Neural Networks}} and {{Tabular Data}}: {{A Survey}}.
\newblock \emph{IEEE Transactions on Neural Networks and Learning Systems},
  pages 1--21, 2022.
\newblock ISSN 2162-237X, 2162-2388.
\newblock \doi{10.1109/TNNLS.2022.3229161}.

\bibitem[Cohen et~al.(2017)Cohen, Afshar, Tapson, and {van
  Schaik}]{cohenEMNISTExtendingMNIST2017}
Gregory Cohen, Saeed Afshar, Jonathan Tapson, and Andr{\'e} {van Schaik}.
\newblock {{EMNIST}}: {{Extending MNIST}} to handwritten letters.
\newblock In \emph{2017 {{International Joint Conference}} on {{Neural
  Networks}} ({{IJCNN}})}, pages 2921--2926, May 2017.
\newblock \doi{10.1109/IJCNN.2017.7966217}.

\bibitem[Cover(1999)]{cover1999elements}
Thomas~M Cover.
\newblock \emph{Elements of information theory}.
\newblock John Wiley \& Sons, 1999.

\bibitem[Darwiche and Marquis(2002)]{darwicheKnowledgeCompilationMap2002}
A.~Darwiche and P.~Marquis.
\newblock A {{Knowledge Compilation Map}}.
\newblock \emph{Journal of Artificial Intelligence Research}, 17:\penalty0
  229--264, September 2002.
\newblock ISSN 1076-9757.
\newblock \doi{10.1613/jair.989}.

\bibitem[De~Raedt and Kimmig(2015)]{de2015probabilistic}
Luc De~Raedt and Angelika Kimmig.
\newblock Probabilistic (logic) programming concepts.
\newblock \emph{Machine Learning}, 100:\penalty0 5--47, 2015.

\bibitem[De~Raedt et~al.(2007)De~Raedt, Kimmig, and
  Toivonen]{deraedtProbLogProbabilisticProlog2007}
Luc De~Raedt, Angelika Kimmig, and Hannu Toivonen.
\newblock {{ProbLog}}: {{A Probabilistic Prolog}} and {{Its Application}} in
  {{Link Discovery}}.
\newblock In \emph{Proceedings of 20th {{International Joint Conference}} on
  {{Artificial Intelligence}}}, volume~7, pages 2468--2473. AAAI Press, 2007.

\bibitem[Dietterich(1980)]{dietterichApplyingGeneralInduction1980}
Thomas~G. Dietterich.
\newblock Applying general induction methods to the card game eleusis.
\newblock In \emph{Proceedings of the {{First AAAI Conference}} on {{Artificial
  Intelligence}}}, {{AAAI}}'80, pages 218--220, Stanford, California, August
  1980. AAAI Press.

\bibitem[Dietterich and
  Michalski(1981)]{dietterichInductiveLearningStructural1981}
Thomas~G. Dietterich and Ryszard~S. Michalski.
\newblock Inductive learning of structural descriptions: {{Evaluation}}
  criteria and comparative review of selected methods.
\newblock \emph{Artificial Intelligence}, 16\penalty0 (3):\penalty0 257--294,
  July 1981.
\newblock ISSN 0004-3702.
\newblock \doi{10.1016/0004-3702(81)90002-3}.

\bibitem[Ellis et~al.(2018)Ellis, Morales, {Sabl{\'e}-Meyer}, {Solar-Lezama},
  and Tenenbaum]{ellisLearningLibrariesSubroutines2018}
Kevin Ellis, Lucas Morales, Mathias {Sabl{\'e}-Meyer}, Armando {Solar-Lezama},
  and Josh Tenenbaum.
\newblock Learning {{Libraries}} of {{Subroutines}} for {{Neurally}}-- {{Guided
  Bayesian Program Induction}}.
\newblock In \emph{Advances in {{Neural Information Processing Systems}}},
  volume~31, pages 7816--7826. Curran Associates, Inc., 2018.

\bibitem[Ellis et~al.(2019)Ellis, Nye, Pu, Sosa, Tenenbaum, and
  {Solar-Lezama}]{ellisWriteExecuteAssess2019}
Kevin Ellis, Maxwell Nye, Yewen Pu, Felix Sosa, Josh Tenenbaum, and Armando
  {Solar-Lezama}.
\newblock Write, {{Execute}}, {{Assess}}: {{Program Synthesis}} with a
  {{REPL}}.
\newblock In \emph{Advances in {{Neural Information Processing Systems}}},
  volume~32, pages 9169--9178. Curran Associates, Inc., 2019.

\bibitem[Evans and Grefenstette(2018)]{evansLearningExplanatoryRules2018}
Richard Evans and Edward Grefenstette.
\newblock Learning {{Explanatory Rules}} from {{Noisy Data}}.
\newblock \emph{Journal of Artificial Intelligence Research}, 61:\penalty0
  1--64, January 2018.
\newblock ISSN 1076-9757.
\newblock \doi{10.1613/jair.5714}.

\bibitem[Fayyad and
  Irani(1993)]{fayyadMultiIntervalDiscretizationContinuousValued1993}
Usama Fayyad and Keki Irani.
\newblock Multi-{{Interval Discretization}} of {{Continuous-Valued Attributes}}
  for {{Classification Learning}}.
\newblock \emph{International Joint Conference of Artificial Intelligence},
  1993.

\bibitem[Fierens et~al.(2015)Fierens, Broeck, Renkens, Shterionov, Gutmann,
  Thon, Janssens, and De~Raedt]{fierensInferenceLearningProbabilistic2015}
Daan Fierens, Guy Van~Den Broeck, Joris Renkens, Dimitar Shterionov, Bernd
  Gutmann, Ingo Thon, Gerda Janssens, and Luc De~Raedt.
\newblock Inference and learning in probabilistic logic programs using weighted
  {{Boolean}} formulas.
\newblock \emph{Theory and Practice of Logic Programming}, 15\penalty0
  (3):\penalty0 358--401, May 2015.
\newblock ISSN 1471-0684, 1475-3081.
\newblock \doi{10.1017/S1471068414000076}.

\bibitem[Flach(1994)]{flachSimplyLogicalIntelligent1994}
Peter~A. Flach.
\newblock \emph{Simply Logical: Intelligent Reasoning by Example}.
\newblock Wiley Professional Computing. Wiley, Chichester ; New York, 1994.
\newblock ISBN 978-0-471-94152-1 978-0-471-94153-8 978-0-471-94215-3.

\bibitem[Frosst and Hinton(2017)]{frosstDistillingNeuralNetwork2017a}
Nicholas Frosst and Geoffrey Hinton.
\newblock Distilling a {{Neural Network Into}} a {{Soft Decision Tree}}.
\newblock \emph{Comprehensibility and Explanation in AI and ML (CEX), AI* IA},
  2017.

\bibitem[Greff et~al.(2020)Greff, {van Steenkiste}, and
  Schmidhuber]{greffBindingProblemArtificial2020}
Klaus Greff, Sjoerd {van Steenkiste}, and J{\"u}rgen Schmidhuber.
\newblock On the {{Binding Problem}} in {{Artificial Neural Networks}},
  December 2020.

\bibitem[Henle(2021)]{henleChangingRulesLegacy2021}
Jim Henle.
\newblock Changing the {{Rules}}: {{The Legacy}} of {{Robert Abbott}}.
\newblock \emph{The Mathematical Intelligencer}, 43\penalty0 (4):\penalty0
  76--81, December 2021.
\newblock ISSN 1866-7414.
\newblock \doi{10.1007/s00283-021-10116-3}.

\bibitem[{\.I}rsoy et~al.(2012){\.I}rsoy, Y{\i}ld{\i}z, and
  Alpayd{\i}n]{irsoySoftDecisionTrees2012}
Ozan {\.I}rsoy, Olcay~Taner Y{\i}ld{\i}z, and Ethem Alpayd{\i}n.
\newblock Soft decision trees.
\newblock In \emph{Proceedings of the 21st {{International Conference}} on
  {{Pattern Recognition}}}, pages 1819--1822, November 2012.

\bibitem[King and Zeng(2001)]{kingLogisticRegressionRare2001}
Gary King and Langche Zeng.
\newblock Logistic {{Regression}} in {{Rare Events Data}}.
\newblock \emph{Political Analysis}, 9\penalty0 (2):\penalty0 137--163, January
  2001.
\newblock ISSN 1047-1987, 1476-4989.
\newblock \doi{10.1093/oxfordjournals.pan.a004868}.

\bibitem[Lecun et~al.(1998)Lecun, Bottou, Bengio, and
  Haffner]{lecunGradientbasedLearningApplied1998}
Y.~Lecun, L.~Bottou, Y.~Bengio, and P.~Haffner.
\newblock Gradient-based learning applied to document recognition.
\newblock \emph{Proceedings of the IEEE}, 86\penalty0 (11):\penalty0
  2278--2324, November 1998.
\newblock ISSN 1558-2256.
\newblock \doi{10.1109/5.726791}.

\bibitem[Lu et~al.(2019)Lu, Mao, Tenenbaum, and
  Wu]{luNeurallyGuidedStructureInference2019}
Sidi Lu, Jiayuan Mao, Joshua Tenenbaum, and Jiajun Wu.
\newblock Neurally-{{Guided Structure Inference}}.
\newblock In \emph{Proceedings of the 36th {{International Conference}} on
  {{Machine Learning}}}, pages 4144--4153. PMLR, May 2019.

\bibitem[Luo et~al.(2021)Luo, Cheng, Yu, and Yi]{luoSDTRSoftDecision2021}
Haoran Luo, Fan Cheng, Heng Yu, and Yuqi Yi.
\newblock {{SDTR}}: {{Soft Decision Tree Regressor}} for {{Tabular Data}}.
\newblock \emph{IEEE Access}, 9:\penalty0 55999--56011, 2021.
\newblock ISSN 2169-3536.
\newblock \doi{10.1109/ACCESS.2021.3070575}.

\bibitem[Maene et~al.(2025)Maene, Derkinderen, and Zuidberg
  Dos~Martires]{maene2025klay}
Jaron Maene, Vincent Derkinderen, and Pedro Zuidberg Dos~Martires.
\newblock {{KLay}}: {{Accelerating}} arithmetic circuits for neurosymbolic
  {{AI}}.
\newblock In \emph{Proceedings of the {{International Conference}} on
  {{Learning Representations}}}, 2025.

\bibitem[Manhaeve et~al.(2018)Manhaeve, Dumancic, Kimmig, Demeester, and
  De~Raedt]{manhaeveDeepProbLogNeuralProbabilistic2018}
Robin Manhaeve, Sebastijan Dumancic, Angelika Kimmig, Thomas Demeester, and Luc
  De~Raedt.
\newblock {{DeepProbLog}}: {{Neural Probabilistic Logic Programming}}.
\newblock In S~Bengio, H~Wallach, H~Larochelle, K~Grauman, N~{Cesa-Bianchi},
  and R~Garnett, editors, \emph{Advances in {{Neural Information Processing
  Systems}}}, volume~31, pages 3753--3763, Montr{\'e}al, Canada, 2018. Curran
  Associates, Inc.

\bibitem[Manhaeve et~al.(2021)Manhaeve, Duman{\v c}i{\'c}, Kimmig, Demeester,
  and De~Raedt]{manhaeveNeuralProbabilisticLogic2021}
Robin Manhaeve, Sebastijan Duman{\v c}i{\'c}, Angelika Kimmig, Thomas
  Demeester, and Luc De~Raedt.
\newblock Neural probabilistic logic programming in {{DeepProbLog}}.
\newblock \emph{Artificial Intelligence}, 298:\penalty0 103504, September 2021.
\newblock ISSN 0004-3702.
\newblock \doi{10.1016/j.artint.2021.103504}.

\bibitem[Marra and Ku{\v z}elka(2021)]{marraNeuralMarkovLogic2021}
Giuseppe Marra and Ond{\v r}ej Ku{\v z}elka.
\newblock Neural markov logic networks.
\newblock In \emph{Proceedings of the {{Thirty-Seventh Conference}} on
  {{Uncertainty}} in {{Artificial Intelligence}}}, pages 908--917. PMLR,
  December 2021.

\bibitem[Marra et~al.(2024)Marra, Duman{\v c}i{\'c}, Manhaeve, and
  De~Raedt]{marraStatisticalRelationalNeurosymbolic2024}
Giuseppe Marra, Sebastijan Duman{\v c}i{\'c}, Robin Manhaeve, and Luc De~Raedt.
\newblock From statistical relational to neurosymbolic artificial intelligence:
  {{A}} survey.
\newblock \emph{Artificial Intelligence}, 328:\penalty0 104062, March 2024.
\newblock ISSN 0004-3702.
\newblock \doi{10.1016/j.artint.2023.104062}.

\bibitem[Marton et~al.(2024)Marton, L{\"u}dtke, Bartelt, and
  Stuckenschmidt]{martonGradTreeLearningAxisAligned2024a}
Sascha Marton, Stefan L{\"u}dtke, Christian Bartelt, and Heiner Stuckenschmidt.
\newblock {{GradTree}}: {{Learning Axis-Aligned Decision Trees}} with
  {{Gradient Descent}}.
\newblock \emph{Proceedings of the AAAI Conference on Artificial Intelligence},
  38\penalty0 (13):\penalty0 14323--14331, March 2024.
\newblock ISSN 2374-3468.
\newblock \doi{10.1609/aaai.v38i13.29345}.

\bibitem[Mitchell(2013)]{mitchellMachineLearning2013}
Tom~M. Mitchell.
\newblock \emph{Machine Learning}.
\newblock {{McGraw-Hill}} Series in {{Computer Science}}. McGraw-Hill, New
  York, nachdr. edition, 2013.
\newblock ISBN 978-0-07-042807-2 978-0-07-115467-3.

\bibitem[Pedregosa et~al.(2011)Pedregosa, Varoquaux, Gramfort, Michel, Thirion,
  Grisel, Blondel, Prettenhofer, Weiss, Dubourg, Vanderplas, Passos,
  Cournapeau, Brucher, Perrot, and Duchesnay]{scikit-learn}
F.~Pedregosa, G.~Varoquaux, A.~Gramfort, V.~Michel, B.~Thirion, O.~Grisel,
  M.~Blondel, P.~Prettenhofer, R.~Weiss, V.~Dubourg, J.~Vanderplas, A.~Passos,
  D.~Cournapeau, M.~Brucher, M.~Perrot, and E.~Duchesnay.
\newblock Scikit-learn: Machine learning in {P}ython.
\newblock \emph{Journal of Machine Learning Research}, 12:\penalty0 2825--2830,
  2011.

\bibitem[Quinlan(1986)]{quinlanInductionDecisionTrees1986}
J.~R. Quinlan.
\newblock Induction of decision trees.
\newblock \emph{Machine Learning}, 1\penalty0 (1):\penalty0 81--106, March
  1986.
\newblock ISSN 1573-0565.
\newblock \doi{10.1007/BF00116251}.

\bibitem[Quinlan(1990)]{quinlan1990probabilistic}
John~Ross Quinlan.
\newblock Probabilistic decision trees.
\newblock In \emph{Machine learning}, pages 140--152. Elsevier, 1990.

\bibitem[Si et~al.(2019)Si, Raghothaman, Heo, and
  Naik]{siSynthesizingDatalogPrograms2019}
Xujie Si, Mukund Raghothaman, Kihong Heo, and Mayur Naik.
\newblock Synthesizing {{Datalog Programs}} using {{Numerical Relaxation}}.
\newblock In \emph{Proceedings of the {{Twenty-Eighth International Joint
  Conference}} on {{Artificial Intelligence}}}, pages 6117--6124, Macao, China,
  August 2019. International Joint Conferences on Artificial Intelligence
  Organization.
\newblock ISBN 978-0-9992411-4-1.
\newblock \doi{10.24963/ijcai.2019/847}.

\bibitem[Tanno et~al.(2019)Tanno, Arulkumaran, Alexander, Criminisi, and
  Nori]{tannoAdaptiveNeuralTrees2019}
Ryutaro Tanno, Kai Arulkumaran, Daniel Alexander, Antonio Criminisi, and Aditya
  Nori.
\newblock Adaptive {{Neural Trees}}.
\newblock In \emph{Proceedings of the 36th {{International Conference}} on
  {{Machine Learning}}}, pages 6166--6175. PMLR, May 2019.

\bibitem[Wan et~al.(2021)Wan, Dunlap, Ho, Yin, Lee, Jin, Petryk, Bargal, and
  Gonzalez]{wanNBDTNeuralBackedDecision2021}
Alvin Wan, Lisa Dunlap, Daniel Ho, Jihan Yin, Scott Lee, Henry Jin, Suzanne
  Petryk, Sarah~Adel Bargal, and Joseph~E. Gonzalez.
\newblock {{NBDT}}: {{Neural-Backed Decision Trees}}, January 2021.

\bibitem[Yang et~al.(2018)Yang, Morillo, and
  Hospedales]{yangDeepNeuralDecision2018}
Yongxin Yang, Irene~Garcia Morillo, and Timothy~M. Hospedales.
\newblock Deep {{Neural Decision Trees}}.
\newblock 2018.
\newblock \doi{10.48550/ARXIV.1806.06988}.

\bibitem[Yang et~al.(2020)Yang, Ishay, and Lee]{yangNeurASPEmbracingNeural2020}
Zhun Yang, Adam Ishay, and Joohyung Lee.
\newblock {{NeurASP}}: {{Embracing Neural Networks}} into {{Answer Set
  Programming}}.
\newblock In \emph{Proceedings of the 29th {{International Joint Conference}}
  on {{Artificial Intelligence}}}, pages 1755--1762, Yokohama, Japan, July
  2020. International Joint Conferences on Artificial Intelligence
  Organization.
\newblock ISBN 978-0-9992411-6-5.
\newblock \doi{10.24963/ijcai.2020/243}.

\bibitem[Yu et~al.(2023)Yu, Yang, Liu, Wang, and
  Pan]{yuSurveyNeuralsymbolicLearning2023}
Dongran Yu, Bo~Yang, Dayou Liu, Hui Wang, and Shirui Pan.
\newblock A survey on neural-symbolic learning systems.
\newblock \emph{Neural Networks}, 166:\penalty0 105--126, September 2023.
\newblock ISSN 0893-6080.
\newblock \doi{10.1016/j.neunet.2023.06.028}.

\end{thebibliography}

\appendix




\section{Eleusis Experiment: Hidden Concepts}

\label{sec:appendixRules}
\label{app:Rules}

\begin{table}[h!]
    \caption{\label{tab:eleusisRules} Table shows all concepts that we used across the experiment.}
    \footnotesize
  \begin{tabular}{rl}
      \\
      Hidden Order Simple & 
      $
      \begin{aligned}
      & \left[ (Card_0) < rank(Card_1) \right]
      \\
      \land &  \left[ suit(Card_0) < suit(Card_1) \right]          
      \end{aligned}
      $
      \\[2em]

      Hidden Modulo Simple & 
      $
      \begin{aligned}
      &\left[ (rank(Card_0) + 1) \mod |Ranks| = rank(Card_1) \right]
      \\
      \lor  & \left[ (suit(Card_0) + 1) \mod 4  = suit(Card_1) \right]           
      \end{aligned}
      $
      \\[2em]

      Color Parity & 
      $
      \begin{aligned}
      &\bigl[  (parity(Card_{0}) \mod 2 = 1 )   \land (color(Card_{1}) = black\bigr]
      \\ \lor
      & \bigl[ (parity(Card_0) \mod 2 = 0 ) \land (color(Card_1) = red) \bigr]          
      \end{aligned}
       $
      \\[2em]

      Alternating Faces &
      $
      \begin{aligned}
      \phantom{\lor} &face(Card_0) \not = face(Card_1)
      \end{aligned}
      $
      \\[2em]
      
      Alternating Parity &
      $
      \begin{aligned}
      \phantom{\lor} &parity(Card_0) \not = parity(Card_1)
      \end{aligned}
      $
      \\[2em]

      Increase Suits & 
      $
      \begin{aligned}
      \phantom{\lor} &(suit(Card1) + 1 \mod 4) == suit(Card2)
      \end{aligned}
      $
      \\[2em]

      Suit Order & 
      $
      \begin{aligned}
      \phantom{\lor} &suit(card_0) < suit(card_1)
      \end{aligned}
      $
      \\[2em]

      Rank Order & 
      $
      \begin{aligned}
      \phantom{\lor} &rank(card_0) < rank(card_1)
      \end{aligned}
      $
      \\[2em]
  \end{tabular}
\end{table}



\clearpage

\section{Eleusis Experiment: Neural Facts and Attributes}

\FloatBarrier

\begin{table}[h!]
    \centering
    \caption{We list all different neural predicates and facts that are used in the experiment. Our Neural facts are used directly as tests in \neuralPredicate{} and measure whether a relationship holds. Our neural predicates are used to implement tests that use formal logic in \optimalPool{}, \badPool{} and \generalPool{}. The rules are defined in \autoref{fig:appendixEleusisBK}.}
    \label{tab:neural-facts-and-predicates}
    \begin{tabular}{lc}
        \midrule
        \multicolumn{2}{l}{Neural Facts}\\
        \midrule
        $nn(\ImSuit{1}, \ImSuit{1}) :: rel\_suit(\ImSuit{1}, \ImSuit{2})$ & \\
        $nn(ImRank1, ImRank2) :: rel\_suit(ImRank1, ImRank2)$ &  \\
        \midrule
        \multicolumn{2}{l}{Neural Predicates}\\
        \midrule
        $nn(\ImSuit{}, \SymSuit{}) :: suit(\ImSuit{}, \SymSuit{}), \quad \SymSuit{}\in [1,4]$ &  \\
        $nn(\ImRank{}, \SymRank{}) :: rank(\ImRank{}, \SymRank{}), \quad \SymRank{} \in [1,10]$ & \\
        $nn(\ImSuit{}, \Bool{}) :: suit\_attribute(\ImSuit{}, \Bool{})$ & \\
        $nn(\ImRank{}, \Bool{}) :: rank\_attribute(\ImRank{}, \Bool{})$ & \\
    \end{tabular}
\end{table}

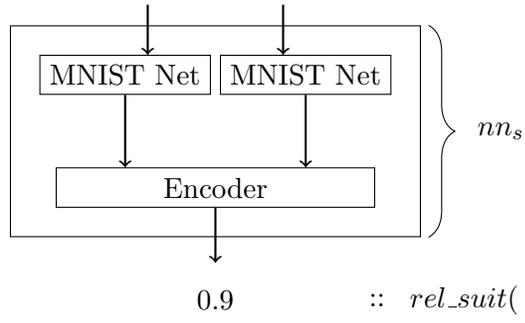
\begin{figure}[h]
    \centering

\begin{tikzpicture}
    \pgfdeclarelayer{foreground}
    \pgfdeclarelayer{background}
    \pgfsetlayers{background, foreground}

    \begin{pgfonlayer}{foreground}

      \path[] (0,1) node[anchor=east] (help-1) {$nn_s($}%
      ++ (0.65,0) node[anchor=center] (imsuit1) {$\ImSuit{1}$}%
         ++ (0.9,0) node[anchor=center, below] {,}%
         ++ (0.9,0) node[anchor=center] (imsuit2) {$\ImSuit{2}$}%
         ++ (0.65,0) node[anchor=west]  {$)$}%
         ++ (0.6,0) node[anchor=center] (middle) {$::$}%
         ++ (0.3,0) node[anchor=west] {$rel\_suit($}%
         ++ (2.4,0) node[anchor=center] {$\ImSuit{1}$}%
         ++ (0.9,0) node[anchor=center, below] {,}%
         ++ (0.9,0) node[anchor=center] {$\ImSuit{2}$}%
         ++ (0.65,0) node[anchor=west] (help-2) {$)$};

       \node[fit=(help-1) (help-2)] (neural-pred-suit) {};

      \node[draw] at ($(imsuit1) + (-0.3,-1.5)$) (decoder1) {MNIST Net};
      \node[draw] at ($(imsuit2) + ( 0.3,-1.5)$) (decoder2) {MNIST Net};

      \node[draw, minimum width=11em] (encoder) at ($(decoder1)!.5!(decoder2) + (0,-1.5)$) {Encoder};

      \draw[->, thick] ($(imsuit1.south) - (0,0.3)$) -- (imsuit1 |- decoder1.north);
      \draw[->, thick] ($(imsuit2.south) - (0,0.3)$) -- (imsuit2 |- decoder2.north);
      \draw[->, thick] (decoder1.south) -- ($(encoder.north -| decoder1.south)$);
      \draw[->, thick] (decoder2.south) -- ($(encoder.north -| decoder2.south)$);

      \coordinate (help) at (encoder -| middle);
      \path[]  ($(help) + (0,-1.5)$) node[anchor=center] (middle2) {$::$}%
         ++ (0.3,0) node[anchor=west] {$rel\_suit($}%
         ++ (2.4,0) node[anchor=center] {$\ImSuit{1}$}%
         ++ (0.9,0) node[anchor=center, below] {,}%
         ++ (0.9,0) node[anchor=center] {$\ImSuit{2}$}%
         ++ (0.65,0) node[anchor=west] (help-2) {$)$};

         \node at (encoder |- middle2) (prob) {0.9};
         \draw[->, thick] (encoder) -- ($(prob) + (0,0.5)$);

         \node[fit=(decoder1) (encoder) (decoder2), draw, inner sep=1em] (nn) {};

         \draw[decorate, decoration={brace, amplitude=10}] ($(nn.east |- nn.north) + (0.1,0)$) -- ($(nn.east |- nn.south) + (0.1,0)$) node [midway, xshift=2.5em] {$nn_s$};
    \end{pgfonlayer}

    \begin{pgfonlayer}{background}
    \end{pgfonlayer}

  \end{tikzpicture}
    \caption{\label{fig:eleusisNeuralAtoms} Figure shows how we define neural atoms in \neuralPredicate{}. A neural atom $rel_suit$ can learn relationships through a \ac{NN} that receives both images as input. In our setting we define neural atoms only on both suits and ranks.}
\end{figure}





\FloatBarrier

\clearpage

\section{Eleusis Experiment: Defined Tests}
\label{fig:appendixEleusisBK}

{\small
	\begin{align*}
	    alternate\_attr\_rank(\ImRank{1}, \ImRank{2}) &\leftarrow%
	    rank\_attribute(\ImRank{1}, \SymRank{1}) ,\\
        &\phantom{\leftarrow}\quad rank\_attribute(\ImRank{2}, \SymRank{2}) ,\\
							  &\phantom{\leftarrow}\quad \SymRank{1} \not= \SymRank{2}\\[0.3em]
	    equal\_ranks(\ImRank{1}, \ImRank{2}) &\leftarrow%
	     rank(\ImRank{1}, \SymRank{1}) ,\\ 
        &\phantom{\leftarrow}\quad rank(\ImRank{2}, \SymRank{2}),\\
						 &\phantom{\leftarrow}\quad\SymRank{1} \not= \SymRank{2} \\[0.3em]
	    modulo\_rank(\ImRank{1}, \ImRank{2}) &\leftarrow%
	    rank(\ImRank{1}, \SymRank{1}),\\ 
        &\phantom{\leftarrow}\quad rank(\ImRank{2}, \SymRank{2}) ,\\
						 &\phantom{\leftarrow}\quad ((\SymRank{1} + 1) \mod 10) =:= \SymRank{2}\\
	    increment\_rank(\ImRank{1}, \ImRank{2}) &\leftarrow%
	    rank(\ImRank{1}, \SymRank{1}),\\
        &\phantom{\leftarrow}\quad rank(\ImRank{2}, \SymRank{2}) ,\\
						    &\phantom{\leftarrow}\quad (\SymRank{1} + 1) =:= \SymRank{2}\\[0.3em]
	    gt\_rank(\ImRank{1}, \ImRank{2}) &\leftarrow%
	    rank(\ImRank{1}, \SymRank{1}) ,
        \\&\phantom{\leftarrow}\quad rank(\ImRank{2}, \SymRank{2}) ,\\
					     &\phantom{\leftarrow}\quad \SymRank{1} < \SymRank{2}\\[0.3em]
	    eq\_rank\_attrs(\ImRank{1}, \ImRank{2}) &\leftarrow%
	    rank_{\theta_1}(\ImRank{1}, \SymRank{1}),\\
        &\phantom{\leftarrow}\quad rank_{\theta_2}(\ImRank{2}, \SymRank{2}) ,\\
						    &\phantom{\leftarrow}\quad \SymRank{1} = \SymRank{2}\\[0.3em]
	    alternate\_attr\_suit(\ImSuit{1}, \ImSuit{2}) &\leftarrow%
	    suit\_attribute(\ImSuit{1}, \SymSuit{1}),\\
        &\phantom{\leftarrow}\quad suit\_attribute(\ImSuit{2}, \SymSuit{2}) ,\\
							  &\phantom{\leftarrow}\quad \SymSuit{1} \not= \SymSuit{2}\\[0.3em]
	    equal\_suits(\ImSuit{1}, \ImSuit{2}) &\leftarrow%
	    suit(\ImSuit{1}, \SymSuit{1}),\\
        &\phantom{\leftarrow}\quad suit(\ImSuit{2}, \SymSuit{2}) ,\\
						 &\phantom{\leftarrow}\quad \SymSuit{1} \not= \SymSuit{2}\\[0.3em]
	    modulo\_suit(\ImSuit{1}, \ImSuit{2}) &\leftarrow%
	    suit(\ImSuit{1}, \SymSuit{1}),\\
        &\phantom{\leftarrow}\quad suit(\ImSuit{2}, \SymSuit{2}) ,\\
						 &\phantom{\leftarrow}\quad ((\SymSuit{1} + 1) \mod 10) =:= \SymSuit{2}\\[0.3em]
	    increment\_suit(\ImSuit{1}, \ImSuit{2}) &\leftarrow%
	    suit(\ImSuit{1}, \SymSuit{1}),\\
        &\phantom{\leftarrow}\quad suit(\ImSuit{2}, \SymSuit{2}) ,\\
						    &\phantom{\leftarrow}\quad (\SymSuit{1} + 1) =:= \SymSuit{2}\\[0.3em]
	    gt\_suit(\ImSuit{1}, \ImSuit{2}) &\leftarrow%
	    suit(\ImSuit{1}, \SymSuit{1}),\\
        &\phantom{\leftarrow}\quad suit(\ImSuit{2}, \SymSuit{2}) ,\\
					     &\phantom{\leftarrow}\quad \SymSuit{1} < \SymSuit{2}\\[0.3em]
	    eq\_suit\_attrs(\ImSuit{1}, \ImSuit{2}) &\leftarrow%
	    suit_{\theta_1}(\ImSuit{1}, \SymSuit{1}),\\
        &\phantom{\leftarrow}\quad suit_{\theta_2}(\ImSuit{2}, \SymSuit{2}) ,\\
						    &\phantom{\leftarrow}\quad \SymSuit{1} = \SymSuit{2}
    \end{align*}
  }

\end{document}